\documentclass[10pt,twocolumn,letterpaper]{article}

\usepackage{cvpr}              
\definecolor{cvprblue}{rgb}{0.21,0.49,0.74}
\usepackage[pagebackref,breaklinks,colorlinks,allcolors=cvprblue]{hyperref}
\usepackage{multirow}
\usepackage{bm}
\usepackage{bbding}
\usepackage{tabularx}
\usepackage{enumitem}
\usepackage[ruled,vlined]{algorithm2e}
\def\paperID{7470} 
\def\confName{CVPR}
\def\confYear{2026}

\title{CountEx: Fine-Grained Counting via Exemplars and Exclusion}

\author{%
  Yifeng Huang\textsuperscript{1}\quad
  Gia Khanh Nguyen\textsuperscript{2}\quad
  Minh Hoai\textsuperscript{2}\\[1ex]
  \textsuperscript{1}Department of Computer Science, Stony Brook University, Stony Brook, NY, USA \\
  \textsuperscript{2}Australian Institute for Machine Learning, Adelaide University, SA, Australia \\[0.5ex]
}

\begin{document}
\def\mA{\mathcal{A}}
\def\mB{\mathcal{B}}
\def\mC{\mathcal{C}}
\def\mD{\mathcal{D}}
\def\mE{\mathcal{E}}
\def\mF{\mathcal{F}}
\def\mG{\mathcal{G}}
\def\mH{\mathcal{H}}
\def\mI{\mathcal{I}}
\def\mJ{\mathcal{J}}
\def\mK{\mathcal{K}}
\def\mL{\mathcal{L}}
\def\mM{\mathcal{M}}
\def\mN{\mathcal{N}}
\def\mO{\mathcal{O}}
\def\mP{\mathcal{P}}
\def\mQ{\mathcal{Q}}
\def\mR{\mathcal{R}}
\def\mS{\mathcal{S}}
\def\mT{\mathcal{T}}
\def\mU{\mathcal{U}}
\def\mV{\mathcal{V}}
\def\mW{\mathcal{W}}
\def\mX{\mathcal{X}}
\def\mY{\mathcal{Y}}
\def\mZ{\mathcal{Z}} 

\def\bbN{\mathbb{N}} 
\def\bbR{\mathbb{R}} 
\def\bbP{\mathbb{P}} 
\def\bbQ{\mathbb{Q}} 
\def\bbE{\mathbb{E}}

\def\1n{\mathbf{1}_n}
\def\0{\mathbf{0}}
\def\1{\mathbf{1}}

\def\A{{\bf A}}
\def\B{{\bf B}}
\def\C{{\bf C}}
\def\D{{\bf D}}
\def\E{{\bf E}}
\def\F{{\bf F}}
\def\G{{\bf G}}
\def\H{{\bf H}}
\def\I{{\bf I}}
\def\J{{\bf J}}
\def\K{{\bf K}}
\def\L{{\bf L}}
\def\M{{\bf M}}
\def\N{{\bf N}}
\def\O{{\bf O}}
\def\P{{\bf P}}
\def\Q{{\bf Q}}
\def\R{{\bf R}}
\def\S{{\bf S}}
\def\T{{\bf T}}
\def\U{{\bf U}}
\def\V{{\bf V}}
\def\W{{\bf W}}
\def\X{{\bf X}}
\def\Y{{\bf Y}}
\def\Z{{\bf Z}}

\def\a{{\bf a}}
\def\b{{\bf b}}
\def\c{{\bf c}}
\def\d{{\bf d}}
\def\e{{\bf e}}
\def\f{{\bf f}}
\def\g{{\bf g}}
\def\h{{\bf h}}
\def\i{{\bf i}}
\def\j{{\bf j}}
\def\k{{\bf k}}
\def\l{{\bf l}}
\def\m{{\bf m}}
\def\n{{\bf n}}
\def\o{{\bf o}}
\def\p{{\bf p}}
\def\q{{\bf q}}
\def\r{{\bf r}}
\def\s{{\bf s}}
\def\t{{\bf t}}
\def\u{{\bf u}}
\def\v{{\bf v}}
\def\w{{\bf w}}
\def\x{{\bf x}}
\def\y{{\bf y}}
\def\z{{\bf z}}

\def\balpha{\mbox{\boldmath{$\alpha$}}}
\def\bbeta{\mbox{\boldmath{$\beta$}}}
\def\bdelta{\mbox{\boldmath{$\delta$}}}
\def\bgamma{\mbox{\boldmath{$\gamma$}}}
\def\blambda{\mbox{\boldmath{$\lambda$}}}
\def\bsigma{\mbox{\boldmath{$\sigma$}}}
\def\btheta{\mbox{\boldmath{$\theta$}}}
\def\bomega{\mbox{\boldmath{$\omega$}}}
\def\bxi{\mbox{\boldmath{$\xi$}}}
\def\bnu{\mbox{\boldmath{$\nu$}}}                                  
\def\bphi{\mbox{\boldmath{$\phi$}}}
\def\bmu{\mbox{\boldmath{$\mu$}}}

\def\bDelta{\mbox{\boldmath{$\Delta$}}}
\def\bOmega{\mbox{\boldmath{$\Omega$}}}
\def\bPhi{\mbox{\boldmath{$\Phi$}}}
\def\bLambda{\mbox{\boldmath{$\Lambda$}}}
\def\bSigma{\mbox{\boldmath{$\Sigma$}}}
\def\bGamma{\mbox{\boldmath{$\Gamma$}}}
                                  
\newcommand{\myprob}[1]{\mathop{\mathbb{P}}_{#1}}

\newcommand{\myexp}[1]{\mathop{\mathbb{E}}_{#1}}

\newcommand{\mydelta}[1]{1_{#1}}

\newcommand{\myminimum}[1]{\mathop{\textrm{minimum}}_{#1}}
\newcommand{\mymaximum}[1]{\mathop{\textrm{maximum}}_{#1}}    
\newcommand{\mymin}[1]{\mathop{\textrm{minimize}}_{#1}}
\newcommand{\mymax}[1]{\mathop{\textrm{maximize}}_{#1}}
\newcommand{\mymins}[1]{\mathop{\textrm{min.}}_{#1}}
\newcommand{\mymaxs}[1]{\mathop{\textrm{max.}}_{#1}}  
\newcommand{\myargmin}[1]{\mathop{\textrm{argmin}}_{#1}} 
\newcommand{\myargmax}[1]{\mathop{\textrm{argmax}}_{#1}} 
\newcommand{\myst}{\textrm{s.t. }}

\newcommand{\denselist}{\itemsep -1pt}
\newcommand{\sparselist}{\itemsep 1pt}

\definecolor{pink}{rgb}{0.9,0.5,0.5}
\definecolor{purple}{rgb}{0.5, 0.4, 0.8}   
\definecolor{gray}{rgb}{0.3, 0.3, 0.3}
\definecolor{mygreen}{rgb}{0.2, 0.6, 0.2}

\newcommand{\cyan}[1]{\textcolor{cyan}{#1}}
\newcommand{\blue}[1]{\textcolor{blue}{#1}}
\newcommand{\magenta}[1]{\textcolor{magenta}{#1}}
\newcommand{\pink}[1]{\textcolor{pink}{#1}}
\newcommand{\green}[1]{\textcolor{green}{#1}} 
\newcommand{\gray}[1]{\textcolor{gray}{#1}}    
\newcommand{\mygreen}[1]{\textcolor{mygreen}{#1}}    
\newcommand{\purple}[1]{\textcolor{purple}{#1}}       

\definecolor{greena}{rgb}{0.4, 0.5, 0.1}
\newcommand{\greena}[1]{\textcolor{greena}{#1}}

\definecolor{bluea}{rgb}{0, 0.4, 0.6}
\newcommand{\bluea}[1]{\textcolor{bluea}{#1}}
\definecolor{reda}{rgb}{0.6, 0.2, 0.1}
\newcommand{\reda}[1]{\textcolor{reda}{#1}}

\def\changemargin#1#2{\list{}{\rightmargin#2\leftmargin#1}\item[]}
\let\endchangemargin=\endlist
                                               
\newcommand{\cm}[1]{}

\newcommand{\mhoai}[1]{{\color{magenta}\textbf{[MH: #1]}}}
\newcommand{\yifeng}[1]{{\color{blue}\textbf{[yifeng: #1]}}}
\newcommand{\khanh}[1]{{\color{olive}\textbf{[khanh: #1]}}}
\newcommand{\yftodo}[1]{{\color{blue}$\blacksquare$\textbf{[TODO: #1]}}}

\newcommand{\mtodo}[1]{{\color{red}$\blacksquare$\textbf{[TODO: #1]}}}
\newcommand{\myheading}[1]{\vspace{1ex}\noindent \textbf{#1}}
\newcommand{\htimesw}[2]{\mbox{$#1$$\times$$#2$}}


\newif\ifshowsolution
\showsolutiontrue

\ifshowsolution  
\newcommand{\Solution}[2]{\paragraph{\bf $\bigstar $ SOLUTION:} {\sf #2} }
\newcommand{\Mistake}[2]{\paragraph{\bf $\blacksquare$ COMMON MISTAKE #1:} {\sf #2} \bigskip}
\else
\newcommand{\Solution}[2]{\vspace{#1}}
\fi

\newcommand{\truefalse}{
\begin{enumerate}
	\item True
	\item False
\end{enumerate}
}

\newcommand{\Sref}[1]{Sec.~\ref{#1}}
\newcommand{\Eref}[1]{Eq.~(\ref{#1})}
\newcommand{\Fref}[1]{Fig.~\ref{#1}}
\newcommand{\Tref}[1]{Table~\ref{#1}}

\maketitle
\begin{abstract}

This paper presents CountEx, a discriminative visual counting framework designed to address a key limitation of existing prompt-based methods: the inability to explicitly exclude visually similar distractors. While current approaches allow users to specify what to count via inclusion prompts, they often struggle in cluttered scenes with confusable object categories, leading to ambiguity and overcounting. CountEx enables users to express both inclusion and exclusion intent, specifying what to count and what to ignore, through multimodal prompts including natural language descriptions and optional visual exemplars. At the core of CountEx is a novel Discriminative Query Refinement module, which jointly reasons over inclusion and exclusion cues by first identifying shared visual features, then isolating exclusion-specific patterns, and finally applying selective suppression to refine the counting query. To support systematic evaluation of fine-grained counting methods, we introduce CoCount, a benchmark comprising 1,780 videos and 10,086 annotated frames across 97 category pairs. Experiments show that CountEx achieves substantial improvements over state-of-the-art methods for counting objects from both known and novel categories. The data and code are available at \url{https://github.com/bbvisual/CountEx}.

\end{abstract}    
\section{Introduction}
\label{sec:intro}

Visual object counting is a fundamental task in computer vision, with applications ranging from crowd monitoring to medical imaging, and more recently, a benchmark for evaluating the reasoning abilities of vision-language models. While current models can estimate object counts from textual prompts or annotated exemplars, fine-grained counting in scenes with multiple coexisting object categories remains challenging. Models often misinterpret user intent, overcount distractors, or default to dominant classes, underscoring the need for more precise and controllable mechanisms to specify both what to count and what to ignore.

\begin{figure}[t]
  \centering
    \includegraphics[width=0.45\textwidth]{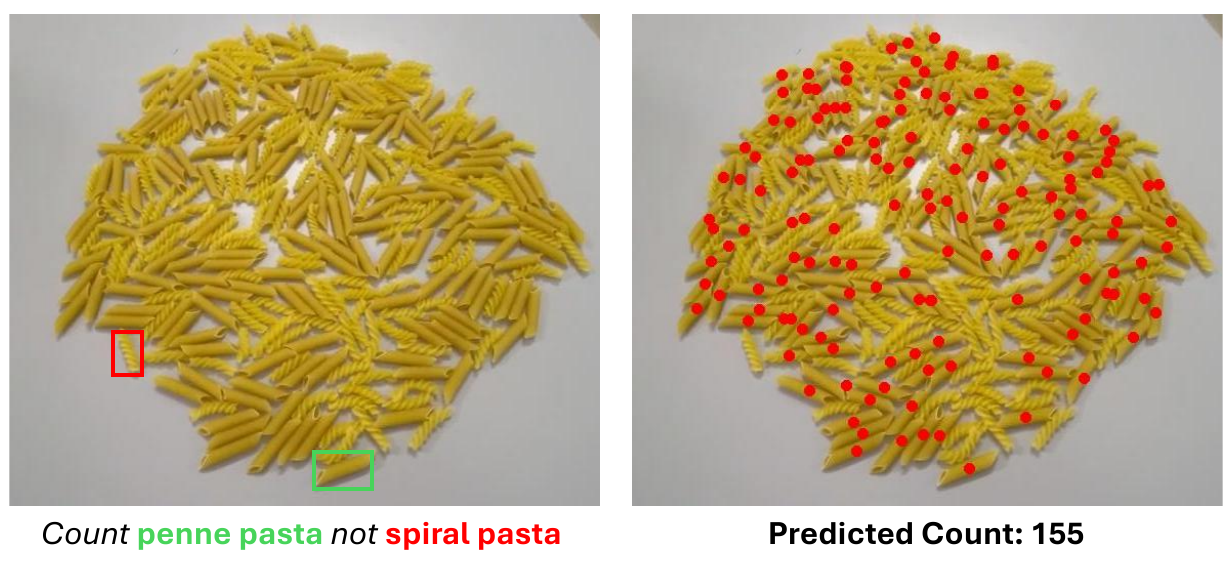}
    \vskip -0.15in
  \caption{Given a cluttered scene containing mutiple object categories, our method allows users to specify both inclusion and exclusion intent, e.g., ``Count penne pasta, not spiral pasta'', via language prompts and optional visual exemplars, enabling more precise and controllable counting.}
    \label{fig:teaser}
    \vskip -0.3in
\end{figure}

In this work, we propose a more expressive interface for visual counting that enables users to explicitly indicate both what they want to count and what they want to exclude, as illustrated in \Fref{fig:teaser}. Users can specify their intent through natural language descriptions alone, or optionally provide exemplar bounding boxes for additional visual guidance. For instance, a user might specify,  ``Count penne pasta, not spiral pasta,'' or ``Count white poker chips, not the blue ones.'' When visual disambiguation is needed, a few example bounding boxes (e.g., three per category) can complement the text descriptions, as shown in \Fref{fig:teaser}. By allowing explicit specification of negative examples, our approach reduces ambiguity and improves counting accuracy in complex scenes with visually similar objects. For brevity, we refer to the objects the user wants to count as {\bf positive}, and those they explicitly want to exclude as {\bf negative}. Note that scenes may also contain {\bf distractor objects}, other categories not mentioned by the user, that should neither be counted nor treated as explicit negatives. 

Although the idea of explicitly specifying what not to count is intuitive and seemingly straightforward, it remains largely unexplored in the literature. While it is reasonable to expect that negative examples can help disambiguate user intent and improve accuracy, it is still unclear how much they help and how best to integrate them into the counting pipeline. Current counting methods are typically rigid in design, accepting user intent for only one object category at a time and lacking the flexibility to interpret both inclusion and exclusion intent. A naive workaround might involve treating positive and negative exemplars separately, computing individual counts or density maps for each, and subtracting one from the other. However, such disjoint treatment ignores the relational context between what should and shouldn't be counted, and often yields poor results.


In this paper, we propose a novel counting architecture that jointly reasons about both positive and negative intent throughout the entire pipeline. Our method encodes multimodal prompts (text and optional exemplars) to extract positive queries and negative-exclusive references, then applies discriminative query refinement to suppress distractor patterns before final counting. By considering inclusion and exclusion simultaneously, the model faithfully captures user intent in scenes with visually similar or co-occurring categories.

To develop and evaluate a method capable of fine-grained counting using both positive and negative exemplars, we introduce a new dataset, referred to as \textbf{CoCount}. Unlike existing counting datasets, which typically feature only a single object category in sufficient quantity, such as~\cite{ranjan2021learning,nguyen2022few,Dai_2024_CVPR}, or multiple categories with too few annotated examples, such as~\cite{d2025just,nguyen2025can}. CoCount contains 1,335 videos and 10,086 annotated frames spanning 97 category pairs, including both \textit{inter-category pairs} (such as coins and paper clips) and \textit{intra-category pairs} (such as straight screws and eye-lag screws). Each image includes two co-occurring object types with ample instances, supporting both training and evaluation of fine-grained counting methods that must reason over inclusion and exclusion intent. This design addresses a key limitation of prior datasets, which often lead models to ignore user prompts and instead count the dominant category due to dataset bias. To the best of our knowledge, CoCount is the first dataset to enable large-scale, exemplar-based fine-grained counting in such complex multi-object scenarios.

In summary, we make three contributions. First, we formulate the task of counting with explicit exclusion cues, allowing users to specify both what to count and ignore. Second, we propose CountEx, a new architecture that jointly reasons over inclusion and exclusion signals, achieving the state-of-the-art performance on multiple benchmarks. Third, we introduce CoCount, a fine-grained benchmark with 10,086 annotated images across 97 category pairs, supporting scalable research on exemplar-based counting.

\section{Related Work}
\label{sec:related_work}


\myheading{Visual counting.} Early visual counting methods were predominantly class-specific, requiring extensive labeled data for each target category~\cite{m_Ranjan-etal-ECCV18,li2018csrnet,zhang2016single,miao2020shallow,jiang2020attention,wang2020DMCount,xu2019learn,hu2020count,bai2020adaptive,liu2020adaptive,song2021rethinking,liu2019point,lian2019density,sam2020locate,liu2019recurrent,laradji2018blobs}. To enable generalization across arbitrary categories, class-agnostic counting  methods emerged, regressing density maps by computing spatial correlations between query images and visual exemplars~\cite{ranjan2021learning,yang2021class,shi2022represent,lu2018class,nguyen2022few,liu2022countr,m_Huang-etal-ICCV23,huang2024point}. Subsequent works introduced prototype matching, attention refinement to improve performance~\cite{djukic2023loca,pelhan2024novel,pelhan2024dave,wang2024vision,Zhiyuan2023safecount,shi2022represent}. Exemplar-free approaches~\cite{ranjan2022exemplar,hobley2022learning} reduce annotation burden but lack control in multi-category scenarios.

\myheading{Counting with language prompts.} Vision-language models enabled text-guided counting, where natural language descriptions replace or complement visual exemplars~\cite{xu2023zero,jiang2023clip,amini2023open,zhu2024zero,countgd,shi2024training,liu2024grounding,zhang2025enhancing,kang2024vlcounter}. ZSC~\cite{xu2023zero} pioneered this by constructing visual prototypes from text, while CLIP-Count~\cite{jiang2023clip} and CounTX~\cite{amini2023open} leveraged CLIP~\cite{radford2021learning} for cross-modal alignment. Recent methods extended GroundingDINO~\cite{liu2024grounding,Dai_2024_CVPR,countgd,wang2025exploring,liu2025countse} . However, these methods share a fundamental limitation: they specify only \emph{what to count} through positive prompts, lacking mechanisms to explicitly exclude visually similar distractors, leading to ambiguity in dense scenes with co-occurring categories.

\myheading{Counting with negative prompts.} While language-guided methods specify target objects through positive prompts, they cannot explicitly exclude visually similar distractors. Recent work~\cite{d2025just} addresses this through test-time adaptation: their method synthesizes images of confounding categories via diffusion models and applies hard negative supervision with empty attention maps, achieving improvements on the LOOKALIKES benchmark. However, this requires per-category optimization and relies on synthetic data quality. In contrast, our approach enables direct negative prompt specification at inference time without additional training or data generation, prioritizing real-time interactivity and user control for immediate disambiguation.

\begin{figure}[t]
  \centering
    \includegraphics[width=0.48\textwidth]{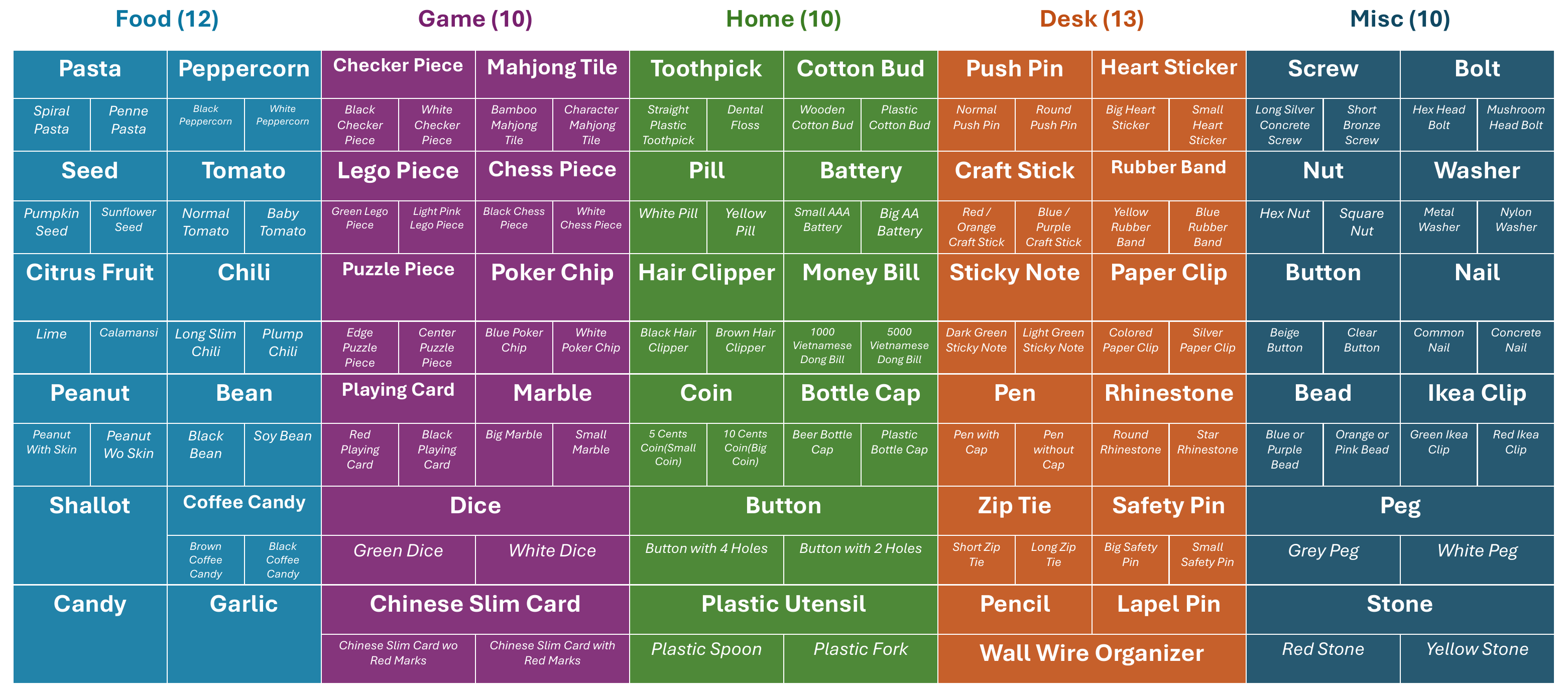}
    \vskip -0.1in
  \caption{Object categories and their variants across five super-categories on CoCount.} 
    \label{fig:dataset_stats}
    \vskip -0.3in
\end{figure}

\section{CoCount Dataset\label{sec:cocount}}

\label{sec:benchmark}

This section describes CoCount, our contributed dataset designed to enable systematic development and evaluation of counting methods with explicit negative prompts. The dataset addresses the need for training and evaluating models that can discriminate fine-grained visual distinctions by explicitly specifying both what to count and what to exclude. CoCount 
comprises 10,086 annotated frames organized into five high-level categories (Food, Game, Home, Desk, Misc), with 97 pairs testing both inter-category discrimination between related objects within the same category and intra-category discrimination between variants of the same object type.

\myheading{Object categories and pairing.}
CoCount is organized into five supercategories(\textit{Food}, \textit{Game}, \textit{Home}, \textit{Desk}, and \textit{Misc}) encompassing 55 distinct object categories, as shown in \Fref{fig:dataset_stats}. Within each category, such as poker chips or pens, we selected up to two visually similar variants that differ in attributes like color, size, or shape. To capture varying levels of visual granularity, we construct evaluation pairs using two strategies. First, we create 50 inter-category pairs by pairing categories within the same supercategory, such as pasta and peppercorn in \textit{Food}, or poker chips and dice in \textit{Game}. Second, we form 50 intra-category pairs, consisting of subcategories within the same category to test fine-grained attribute discrimination, such as black versus white peppercorn, or AA versus AAA batteries.

\myheading{Video recording.} For each of the 100 category pairs described above, we recorded 20 videos per pair, organized as 10 distinct count settings, each filmed twice with objects re-scrambled and reoriented. We did so by starting from the maximum feasible counts for both categories and later decreased counts across five successive groups. Within each group, the two companion videos have closely matched counts (5-10\% difference), while each subsequent group shows a larger decrement. This grouped design intentionally creates near-duplicate scenes with controlled object removals, enabling the data to use as a verification/contrastive resource for training. For every video, we first determined the number of objects from each category and create a scene by spreading those objects across a surface (e.g., table, bed, floor or table cloth). To increase scene diversity and realism, we also added distractor objects not belonging to either target category. Each video was capture using a handheld camera while walking around and zooming in and out the scene to simulate natural viewing conditions.



\myheading{Data split and annotation.} These videos vary in object categories, counts, backgrounds, and layouts. Each video contains multiple frames annotated with two dominant object categories and their counts. This makes the dataset a valuable resource for training and evaluation. To facilitate experimentation, we divide the videos into four disjoint subsets: \textbf{Train}, \textbf{Train2}, \textbf{Val}, and \textbf{Test}, and extract representative frames for annotation. The number of videos, along with additional statistics such as the number of annotated frames and available annotation types for each subset, are summarized in~\Tref{tab:num_videos} and described below.

For the \textbf{Train} set, where dense annotations are needed to train counting models, we adopt a semi-automated annotation pipeline designed to balance quality and efficiency. Specifically, we manually annotate one clear frame per video with a dot at the center of each object instance, along with three exemplar bounding boxes for each of the two categories in the category pair. These annotations are then propagated to the remaining frames using CoTracker3~\cite{karaev2025cotracker3}, an off-the-shelf pixel tracker. After temporal sampling and quality filtering, we retain 7,417 annotated frames for training, covering a wide range of object configurations, occlusion levels, and scale variations.

Due to the labor-intensive nature of dot annotation—even at a rate of one frame per video—we did not annotate the \textbf{Train2} set and ultimately exclude it from the experiments in this paper. Nonetheless, we report its availability for completeness. In future work, this subset could be annotated or used in semi-supervised or weakly supervised settings, as total object counts per video are known and could provide useful weak supervision.

For the \textbf{Val} set, we randomly sample three frames per video (1,335 frames in total) to provide a balanced subset for hyperparameter tuning. For the \textbf{Test} set, we manually select three high-quality frames per video based on clarity (minimal motion blur and good lighting) and visual diversity (viewpoint and layout), resulting in 1,334 test frames after discarding one low-quality example. To support both exemplar-based and text-guided methods, we provide three manually annotated exemplar bounding boxes per object variant in all test frames.

\setlength{\tabcolsep}{2pt}

\begin{table}[t]
\centering
\begin{tabular}{clcccc}
\toprule
 & & \textbf{Train} & \textbf{Train2} & \textbf{Val} & \textbf{Test} \\
\midrule
\multirow{6}{*}{\rotatebox[origin=c]{90}{\textbf{\# videos}}}
 & Food & 92 & 92 & 92 & 92\\
 & Game & 92 & 92 & 92 & 92 \\
 & Home & 95 & 95 & 95 & 95\\
 & Desk & 86 & 86 & 86 & 86 \\
 & Misc & 80 & 80 & 80 & 80\\
 & Total & 445 & 445 & 445 & 445\\
 \midrule 
\multirow{5}{*}{\rotatebox[origin=c]{90}{\textbf{Annotation}}}
& \# annotated frames  & 7417 & 0 & 1335 & 1334 \\
& \# total objects & 1,159K & - & 204K & 204K \\
 & Per-category object counts & \checkmark & \checkmark & \checkmark & \checkmark \\
 & B. boxes for six exemplars & \checkmark & & & \checkmark  \\
 & Dots for all objects & \checkmark & \\
 \bottomrule
\end{tabular}
\vskip -0.1in
\caption{Video counts and annotation across dataset splits. \label{tab:num_videos} }
\vskip -0.3in
\end{table}




\myheading{Supported experiment settings.} The five supercategories enable two complementary evaluation protocols for fine-grained counting. The first is the \textbf{Novel-Category Setting (NC-setting)}, which evaluates a model's ability to generalize to object categories not seen during training. This setting takes advantage of the dataset's supercategory structure: Test-Food, Test-Game, Test-Home, Test-Desk, and Test-Misc. For each test subset, all training data containing the same categories are excluded. For example, when evaluating on Test-Food, the model is trained using Train-Game, Train-Home, Train-Desk, and Train-Misc, while Train-Food is omitted entirely.

The second protocol is the \textbf{Known-Category Setting (KC-setting)}, in which the training set includes categories that also appear in the test set. Note that while object categories may overlap, the test images are extracted from videos that are strictly disjoint from those used to extract the training images. In this setting, a single model is trained on the full training data (i.e., Train-Food, Train-Game, Train-Home, Train-Desk, and Train-Misc) and evaluated across all test subsets. This setting reflects scenarios where some labeled data is available for each target category.



\begin{figure*}[ht]
\centering
    \includegraphics[width=0.8\textwidth]{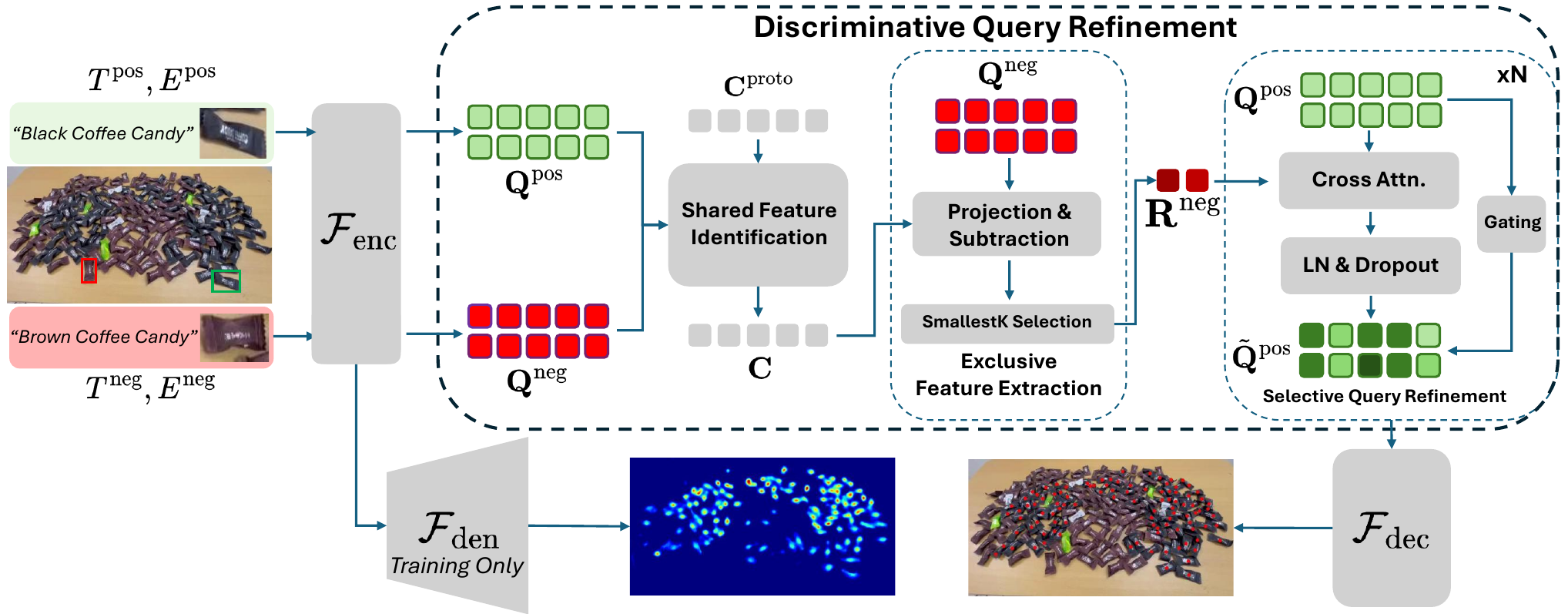}
    \vskip -0.15in
    \caption{Overview of CountEx. Given an image and multimodal prompts (positive and negative text with optional visual exemplars), we encode them into query sets $\mathbf{Q}^{\text{pos}}$ and $\mathbf{Q}^{\text{neg}}$. Our Discriminative Query Refinement (DQR) module consists of three stages: \textbf{(1)} Shared Feature Identification learns prototypes $\mathbf{C}$ capturing common features between both query sets; \textbf{(2)} Exclusive Feature Extraction isolates negative-exclusive patterns $\mathbf{R}^{\text{neg}}$ by projecting $\mathbf{Q}^{\text{neg}}$ onto $\mathbf{C}$ and filtering residuals; \textbf{(3)} Selective Query Refinement produces refined queries $\tilde{\mathbf{Q}}^{\text{pos}}$ by selectively suppressing negative patterns via attention. The refined queries are fed to detection heads for final predictions. An auxiliary density prediction branch provides additional supervision during training.}
\label{fig:pipeline}
\vskip -0.25in
\end{figure*}

\section{CountEx}
\label{sec:approach}



Our task is to develop a model for fine-grained open-vocabulary counting that can leverage explicit negative prompts when available. We formulate the task as follows. Given an image $I$ and a required positive text prompt $T^{\text{pos}}$ describing the objects the user wants to count, the model may also receive any combination of the following optional inputs: positive exemplar bounding boxes $E^{\text{pos}}$, a negative text prompt $T^{\text{neg}}$, and/or negative exemplar bounding boxes $E^{\text{neg}}$. The goal is to predict the total number of instances belonging to the positive object category specified by $T^{\text{pos}}$. The model is expected to flexibly handle varying input configurations, whether none, one, or all optional inputs are provided, with performance depending on the availability and quality of those inputs.




\subsection{Key challenge and approach overview}

A key question is how to effectively use the negative prompt when it is available. This is not as straightforward as it may seem, since the negative prompt is not simply the opposite of the positive one. Naively subtracting one from the other can lead to suboptimal results. For example, in the prompt ``count red apples, not green ones,'' the user intends to exclude green apples, but both red and green apples belong to the same category. Subtracting their embeddings may shift focus toward color differences rather than reinforcing the notion of apples as the relevant object class.

This nuance is important because scenes often contain other, unrelated object types. The model must distinguish between positive, negative, and irrelevant categories, not just contrast subtypes within a category. Designing mechanisms that handle this kind of exclusion without collapsing the broader object context is therefore essential for fine-grained counting.

To address this challenge, we introduce CountEx, which features a novel \textit{Discriminative Query Refinement (DQR)} module. CountEx builds on a query-based detection model~\cite{llmdet}, which begins with an image-and-prompt–conditioned query set $\Q$, predicts detection boxes and scores ${(B_i, s_i)}$, and outputs the total count by enumerating detections with scores above a confidence threshold: $N = \sum_{i=1}^{n} \delta(s_i > \tau)$. CountEx extends this model by encoding the input image along with positive and negative prompts into separate query sets $\mathbf{Q}^{\text{pos}}$ and $\mathbf{Q}^{\text{neg}}$ (\Sref{sec:query_encoding}), and applying DQR (\Sref{sec:dqr}) to refine $\mathbf{Q}^{\text{pos}}$ by isolating negative-exclusive features and suppressing them while preserving shared, category-relevant cues. This enables accurate and discriminative counting in the presence of subtle attribute-level differences. See ~\Fref{fig:pipeline} for an overview.



\subsection{Prompt-Conditioned Query Encoding}
\label{sec:query_encoding}


Our approach is compatible with any open-vocabulary detector that supports multimodal conditioning, such as Grounding DINO~\cite{liu2024grounding} or similar architectures. In this work, we adopt the query encoder from LLMDet~\cite{llmdet}, denoted as $\mathcal{F}_{\text{enc}}$, due to its strong open-vocabulary capabilities. The encoder $\mathcal{F}_{\text{enc}}$ is a transformer encoder-decoder architecture where learnable queries cross-attend to vision and language features to produce a set of $n$ queries encoding potential object instances. To incorporate both text and visual exemplars, we construct multimodal conditioning sequences as follows: text descriptions are encoded via the language encoder to obtain token embeddings. For visual exemplars, each crop $e \in E^{\text{pos}} \cup E^{\text{neg}}$ is encoded by extracting patch features, average-pooling to obtain $\mathbf{v}_e\in\mathbb{R}^{d}$, projecting through a linear layer, and concatenating with text embeddings. We then condition the query encoder separately on the positive and negative prompts $(T_{\text{pos}},E_{\text{pos}})$ and $(T_{\text{neg}},E_{\text{neg}})$ to obtain two query sets: 
\begin{equation}
\mathbf{Q}^{\text{pos}} = \mathcal{F}_{\text{enc}}(I, T^{\text{pos}},E^{\text{pos}}), 
\mathbf{Q}^{\text{neg}} = \mathcal{F}_{\text{enc}}(I, T^{\text{neg}},E^{\text{neg}}), \nonumber 
\end{equation}
where $\mathbf{Q}^{\text{pos}}, \mathbf{Q}^{\text{neg}} \in \mathbb{R}^{n \times d}$. Each query set encodes object candidates aligned with its corresponding prompt, forming the foundation for discriminative refinement. Importantly, by generating separate positive and negative query sets from the same image, we obtain two complementary representations of the scene: one focused on the target objects and the other on visually similar distractors. This dual encoding enables our DQR module to explicitly reason about the distinction between what to count and what to exclude.

\subsection{Discriminative Query Refinement}
\label{sec:dqr}


Given the positive and negative query sets from Sec.~\ref{sec:query_encoding}, we refine $\mathbf{Q}^{\text{pos}}$ using a discriminative process that preserves category-relevant features while selectively suppressing those unique to the negative prompt. Rather than relying on naive subtraction, which can eliminate critical features of the target class, we adopt a three-stage refinement strategy: (1) extract shared features, (2) identify negative-exclusive components, and (3) apply targeted suppression to remove only the truly discriminative negatives.

\subsubsection{Shared Feature Identification}

Given $\mathbf{Q}^{\text{pos}}, \mathbf{Q}^{\text{neg}} \in \mathbb{R}^{n \times d}$, 
this stage identifies shared features by learning $r$ prototypes that capture 
visual attributes common to both positive and negative queries. The intuition 
is that these prototypes should respond equally to queries encoding shared 
attributes such as object shape, texture, or category-level properties. By 
explicitly modeling a shared feature subspace, we can subsequently decompose 
negative queries into components that overlap with positive queries (which 
should be preserved) and components unique to negatives (which should be 
suppressed).

We introduce $r$ learnable prototype embeddings $\mathbf{C}^{\text{proto}} \in 
\mathbb{R}^{r \times d}$, initialized as normalized random vectors. To make 
these prototypes context-aware, we fuse them with information from both query 
sets via cross-attention over the concatenated queries:
\begin{equation}
\mathbf{H} = \textsl{MHA}(\mathbf{C}^{\text{proto}},\,[\mathbf{Q}^{\text{pos}}; \mathbf{Q}^{\text{neg}}],\,[\mathbf{Q}^{\text{pos}}; \mathbf{Q}^{\text{neg}}]),
\end{equation}
\begin{equation}
\mathbf{C} = \textsl{LN}(\textsl{Linear}(\mathbf{H})),
\end{equation}
where \textsl{MHA} denotes the multi-head attention block, \textsl{LN} stands for the layer normalization operation, and \textsl{Linear} refers to the linear layer. The concatenation $[\mathbf{Q}^{\text{pos}}; \mathbf{Q}^{\text{neg}}] \in \mathbb{R}^{2n \times d}$ merges both sets along the sequence dimension, and $\mathbf{C} \in \mathbb{R}^{r \times d}$ represents the shared query prototypes.

Two auxiliary losses guide the learning process. The shareability loss 
encourages each prototype to be similar to at least one query from each set:
\begin{equation}
\mathcal{L}_{\text{share}}
= -\frac{1}{r}\sum_{j=1}^{r}\left[\max_{i}\langle\mathbf{c}_j,\,\mathbf{q}^{\text{pos}}_{i}\rangle
+\max_{i}\langle\mathbf{c}_j,\,\mathbf{q}^{\text{neg}}_{i}\rangle\right], 
\end{equation}
where $\langle\cdot,\cdot\rangle$ denotes cosine similarity and $\mathbf{c}_j$ denotes the $j$-th prototype. This ensures the prototypes capture genuinely shared features rather than category-specific patterns. The diversity loss prevents collapse by encouraging orthogonal prototype embeddings: $\mathcal{L}_{\text{div}}=\|\mathbf{C}^{\text{proto}}{\mathbf{C}^{\text{proto}}}^{\!\top} - \mathbf{I}\|_F^2.$
The total loss for prototype learning is
\begin{align}
\mathcal{L}_{\text{proto}} = \lambda_{\text{share}}\mathcal{L}_{\text{share}} + \lambda_{\text{div}}\mathcal{L}_{\text{div}}. \label{eqn:protoloss}  
\end{align}

\subsubsection{Exclusive Feature Extraction}

Given shared prototypes $\mathbf{C} \in \mathbb{R}^{r \times d}$, this module extracts features unique to the negative prompt through projection-based decomposition followed by exclusivity-based filtering. This two-stage design extracts negative-exclusive patterns while discarding shared features that would harm positive detection if naively subtracted.

We begin by identifying negative queries that are most distant from the 
shared feature space. For each negative query $\mathbf{q}^{\text{neg}}_{i}$, we compute an exclusivity score based on its maximum similarity to any shared prototype:
\begin{equation}
\sigma_i = \max_{j}\,\langle\mathbf{q}^{\text{neg}}_{i},\,\mathbf{c}_{j}\rangle,
\end{equation}
where lower scores indicate further distance from the shared space, suggesting features more exclusive to the negative category. Let $\mI_{\text{excl}}$ denote the set of indices $i$ corresponding to the $m$ smallest $\sigma_i$ values.



For each selected query, we then decompose it into shared and exclusive 
components through subspace projection. The exclusive component is obtained by projecting onto the subspace spanned by $\C$ and extracting the residual:
\begin{equation}
\mathbf{q}^{\text{excl}}_{i} = \mathbf{q}^{\text{neg}}_{i} - \mathbf{q}^{\text{neg}}_{i}\C^{\top}\mathbf{C}, \quad \forall i \in \mathcal{I}_{\text{excl}}.
\end{equation}
This projection removes the shared component, retaining only the 
negative-exclusive residual. Assembling the residuals from selected queries yields the negative-exclusive reference set:
\begin{equation}
\mathbf{R}^{\text{neg}} = \{\mathbf{q}^{\text{excl}}_{i} \mid i \in \mathcal{I}_{\text{excl}}\} \in \mathbb{R}^{m \times d}.
\end{equation}
This two-step design, that first selecting queries with high exclusivity, then extracting their exclusive components, yields a compact set of distinctive negative patterns for subsequent refinement, free from shared features that characterize both categories.

\subsubsection{Selective Query Refinement}
Given negative-exclusive references $\mathbf{R}^{\text{neg}}$ from the previous stage, this module selectively refines positive queries by suppressing negative patterns while preserving discriminative information. As illustrated in \Fref{fig:pipeline}, positive queries $\mathbf{Q}^{\text{pos}}$ are refined through cross-attention to negative-exclusive features: queries attend to $\mathbf{R}^{\text{neg}}$ to identify alignment with distractor patterns, and the attended features are then subtracted via a gated residual connection. Specifically, we apply vanilla cross-attention where $\mathbf{Q}^{\text{pos}}$ serves as queries and $\mathbf{R}^{\text{neg}}$ serves as keys and values, followed by layer normalization and dropout to obtain $\tilde{\mathbf{Q}}^{\text{pos}}$. A learnable gating parameter controls the suppression strength, enabling the model to adaptively balance between aggressive negative filtering and preservation of positive instances. This design follows residual learning principles: queries with low attention to $\mathbf{R}^{\text{neg}}$ pass through largely unchanged, while queries strongly aligned with negative patterns are selectively suppressed. Since $\mathbf{R}^{\text{neg}}$ contains only negative-exclusive features by construction, this refinement avoids removing shared visual attributes and prevents false negatives. The refined queries $\tilde{\mathbf{Q}}^{\text{pos}}$ retain discriminative information about positive instances while being cleaned of distractor patterns. These refined queries are then fed to the dot decoder $\mathcal{F}_{\text{dec}}$ with standard detection heads for classification and box regression to produce final predictions $\{(b_i, s_i)\}_{i=1}^{n}$.

\subsection{Training Objective}
\label{sec:loss}

Following common practice in point-supervised object counting~\cite{Dai_2024_CVPR, countgd, wang2025exploring}, our model is trained end-to-end using a multi-component loss that combines classification, localization, and density prediction terms, along with the prototype learning loss described in \Eref{eqn:protoloss}: $\mathcal{L} = \lambda_{\text{cls}}\mathcal{L}_{\text{cls}} + \mathcal{L}_{\text{loc}} + \lambda_{\text{den}}\mathcal{L}_{\text{den}} + \mathcal{L}_{\text{proto}},$ where $\lambda_{\text{cls}}=5$, $\lambda_{\text{share}}=2$, $\mathcal{L}_{\text{div}}=0.01$ and $\lambda_{\text{den}}=200$.


The Classification Loss applies focal loss over the matched pairs: $\mathcal{L}_{\text{cls}} =\sum_{i=1}^{n} \text{FocalLoss}(s_i, y_i),$ where $s_i$ is the predicted score for bounding box $B_i$, and $y_i$ is the assigned ground-truth label.

The Localization Loss supervises the centers of the matched predicted bounding boxes using the ground-truth point annotations: $\mathcal{L}_{\text{loc}} = \sum_{i=1}^{k} \|\text{center}(B_{i}) - C_i\|_1$, where $B_i$ is the predicted box for the $i$-th query, $C_i$ is the ground-truth point coordinate, $k$ indicates the number of matched queries. 


The Density Prediction Loss provides complementary dense supervision to promote spatially-aware feature learning. We extract text-fused visual tokens from the encoder (prior to query decoding) and pass them through an FPN-based density head to produce a predicted density map $\hat{D}$. For supervision, we construct pseudo ground-truth density maps $D$ from point annotations by placing a 2D Gaussian kernel at each annotated point location. The density prediction is then supervised using mean squared error loss: $\mathcal{L}_{\text{den}} = |\hat{D} - D|_2^2$.




\section{Experiments}
This section describes the experiments conducted to evaluate CountEx across various settings. We benchmark CountEx on CoCount under both novel and known category settings, then assess its generalization on LOOKALIKES~\cite{d2025just}, PairTally~\cite{nguyen2025can}, and FSC-147~\cite{ranjan2021learning}. We also conduct ablations on key components and discuss limitations.

\subsection{Results on CoCount}
We evaluate CountEx on CoCount under two settings: counting novel (NC) and known (KC) categories, as outlined in \Sref{sec:cocount}. We compare against four strong baselines(LLMDet~\cite{llmdet}, CAD-GD~\cite{wang2025exploring}, GroundingREC~\cite{Dai_2024_CVPR}, and CountGD~\cite{countgd}) trained under the same protocols. Following standard practice~\cite{yang2021class, ranjan2021learning, lu2018class, nguyen2022few}, we use Mean Absolute Error (MAE) and Root Mean Squared Error (RMSE) for evaluation. Unlike baselines that only use inclusion prompts, CountEx leverages both inclusion and exclusion cues (textual and visual) for more discriminative counting.

\begin{table}[t]
\centering
\begin{tabular}{lcccc}
\toprule
\multirow{2}{*}{Method} & \multicolumn{2}{c}{NC-setting} & \multicolumn{2}{c}{KC-setting} \\
& MAE & RMSE & MAE & RMSE \\
\midrule
LLMDet~\cite{llmdet} & 33.22 & 47.66 & 16.82 & 29.23 \\
CAD-GD~\cite{wang2025exploring} & 34.08 & 50.39 & 16.00 & 27.52 \\
GroundingREC~\cite{Dai_2024_CVPR} & 29.29 & 42.43 & 17.54 & 27.41 \\
CountGD~\cite{countgd} & 33.78 & 48.29 & 15.55 & 28.32 \\
CountEx (proposed) & \textbf{26.61} & \textbf{38.86} & \textbf{12.72} & \textbf{23.99} \\
\bottomrule
\end{tabular}
\vskip -0.1in
\caption{Performance comparison on the CoCount test set. All methods are trained using the same data under either the known-category (KC) or novel-category (NC) setting. CountEx outperforms all baselines across both settings.} 
\vskip -0.3in

\label{tab:main_results}
\end{table}



\Tref{tab:main_results} presents results on both evaluation protocols. On the task of counting novel object categories (NC-setting), where models train on four categories and test on the held-out fifth to assess zero-shot generalization, our method achieves 26.61 MAE and 38.86 RMSE. Compared to our base architecture LLMDet (33.22 MAE), we achieve a substantial 19.9\% error reduction, demonstrating that discriminative query refinement with explicit negative prompts significantly enhances generalization to novel categories.

On the task of counting objects from known categories (KC-setting), where all five categories are available during training, our method achieves 12.72 MAE and 23.99 RMSE, outperforming the best baseline (CountGD at 15.55 MAE) by 18\%. Notably, while all methods perform better when trained with access to all categories, our method maintains consistent advantages across both settings. The relative improvement over LLMDet is similar in both protocols (19.9\% on NC-setting vs 24.4\% on KC-setting), indicating that our approach benefits both zero-shot and supervised scenarios. Per-category breakdowns for all five NC-setting splits are provided in the supplementary.



\subsection{Evaluation on Other Datasets}
To assess generalization beyond CoCount, we evaluate CountEx and other models trained on the full CoCount data split on the LOOKALIKES benchmark~\cite{d2025just}, a test-only dataset with 1,037 images across 27 fine-grained subcategories. As shown in \Tref{tab:main_exp_lookalikes}, CountEx achieves an MAE of 18.53 under zero-shot transfer without fine-tuning, setting a new state-of-the-art among zero-shot methods. It outperforms {CountGD} by 17.1\% (22.34 $\rightarrow$ 18.53) and GroundingDINO by 45.3\% (33.89 $\rightarrow$ 18.53). For comparison, D'Alessandro et al.~\cite{d2025just} report a lower MAE of 10.00, but their method relies on synthetic data generation and test-time adaptation per category, incurring 5--7 minutes of processing time per class.



\setlength{\tabcolsep}{3pt}
\begin{table}[t]
\centering
\begin{tabular}{lrr}
\toprule
Method & MAE & RMSE \\
\midrule
D'Alessandro et al.~\cite{d2025just} & 10.00 & 16.95 \\
\multicolumn{3}{l}{\colorbox{gray!30}{\it (per category synthetic data generation + adaptation)}} \\
\midrule
\multicolumn{1}{l}{\colorbox{gray!30}{\it Zero-shot - no training/adaptation}} \\
OWLv2~\cite{minderer2023scaling} & 37.25 & 55.01 \\
GroundingDINO~\cite{liu2024grounding} & 33.89 & 59.51 \\
PseCo~\cite{huang2024point} & 59.82 & 72.79 \\
DAVE~\cite{pelhan2024dave} & 56.10 & 73.34 \\
CountGD~\cite{countgd} & 22.34 & 33.90 \\
LLMDet*~\cite{llmdet} & 22.86 & 33.29\\
CAD-GD*~\cite{wang2025exploring} & 28.14 & 41.30 \\
GroundingREC*~\cite{Dai_2024_CVPR} & 27.24 & 41.57\\
CountGD*~\cite{countgd} & 23.55 & 36.89 \\
CountEx* (proposed) & 18.53 & 30.46 \\
\bottomrule
\end{tabular}
\vskip -0.1in
\caption{Results on LOOKALIKES~\cite{d2025just}. * denotes training on 
CoCount before zero-shot evaluation. CountEx achieves best zero-shot 
performance. D'Alessandro et al.~\cite{d2025just} obtain lowest MAE but 
require synthetic data generation and per-category test-time adaptation 
(5--7 minutes per category).}
\vskip -0.3in
\label{tab:main_exp_lookalikes}
\end{table}

\setlength{\tabcolsep}{3pt}
\begin{table}[t]
  \centering
  \begin{tabular}{lrrrr}
  \toprule Model & \multicolumn{2}{c}{MAE $\downarrow$} & \multicolumn{2}{c}{NAE $\downarrow$} \\
\cmidrule(lr){2-3}\cmidrule(lr){4-5} & Inter & Intra & Inter & Intra \\
\midrule 
\multicolumn{5}{l}{\colorbox{gray!30}{\it Pre-trained specialist counting models}} \\
DAVE \cite{pelhan2024dave} & 46.27 & 46.75 & 0.779 & 0.797 \\
GeCo \cite{pelhan2024novel} & 45.05 & 54.80 & 0.777 & 0.935 \\
LoCA \cite{djukic2023loca} & 71.89 & 57.45 & 1.177 & 0.950 \\
FamNet \cite{ranjan2021learning} & 66.97 & 74.75 & 1.363 & 1.440 \\
Count~GD \cite{countgd} & 39.78 & 56.54 & 0.673 & 0.906 \\
Count~GD~(Text) \cite{countgd} & 50.23 & 53.93 & 0.712 & 0.841 \\
LLMDet \cite{llmdet} & 78.72 & 142.08 & 0.661 & 1.060 \\
\midrule
\multicolumn{5}{l}{\colorbox{gray!30}{\it Pre-trained large vision language models}} \\
Ovis2 \cite{lu2024ovis} & 56.87 & 74.24 & 0.711 & 0.736 \\
Qwen2.5-VL \cite{Qwen2.5-VL} & 46.35 & 67.86 & 0.598 & 0.712 \\
LLaMA-3.2 \cite{dubey2024llama} & 49.14 & 58.73 & 0.730 & 0.740 \\
InternVL3\cite{zhu2025internvl3exploringadvancedtraining} & 55.89 &  71.47 &  0.667 &  0.721 \\
\midrule
\multicolumn{5}{l}{\colorbox{gray!30}{\it Models trained on CoCount}} \\
LLMDet~\cite{llmdet} & 18.78 & 21.06 & 0.225 & 0.368\\
CAD-GD~\cite{wang2025exploring} & 19.47 & 16.98 & 0.243 & 0.259\\
GroundingREC~\cite{Dai_2024_CVPR} & 19.71 & 17.91 & 0.260 & 0.349 \\
CountGD~\cite{countgd} & 19.67 & 15.67 & 0.263 & 0.318\\
CountEx (proposed) & 15.61 & 12.57 & 0.197 & 0.229 \\
\bottomrule
\end{tabular}
\vskip -0.1in
\caption{Results on PairTally~\cite{nguyen2025can}. CountEx achieves the best performance on both MAE and NAE, outperforming pre-trained specialist counters, general vision-language models, and models trained on the same CoCount dataset. } 
\vskip -0.3in
\label{tab:pair_tally}
\end{table}

We further evaluate our proposed method on the PairTally benchmark~\cite{nguyen2025can}, comparing it against both specialist counting models and general vision-language models as reported in~\cite{nguyen2025can}. Results are presented in \Tref{tab:pair_tally}, reporting MAE and Normalized Absolute Error (NAE) for both inter-scene and intra-scene settings, following the evaluation protocol in~\cite{nguyen2025can}. CountEx achieves superior performance across all metrics, outperforming not only pre-trained specialist counters and general vision-language models, but also models trained on the same CoCount dataset used for training CountEx.

Another popular benchmark is FSC-147~\cite{ranjan2021learning}, though it primarily contains a single object category per image and therefore offers no opportunity to leverage exclusion prompts. For fair comparison with other methods trained specifically on this dataset, we fine-tune CountEx on the FSC-147 training set and evaluate it using inclusion prompts (from~\cite{amini2023open}) only. On the test set, CountEx achieves an MAE of 8.63, slightly lower than the state-of-the-art CountGD (5.74), but outperforming several recent methods, including DAVE (8.66), CACViT (9.13), LOCA (10.79), CountTR (11.95), VLCounter (17.05), GroundingREC (10.12), and CAD-GD (10.35). Detailed result on FSC-147 are in our supplementray.




\subsection{Ablation Studies}
\myheading{Prompt Combinations.} \Tref{tab:ablation_modality} evaluates performance across different combinations of input modalities on the test data of CoCount. We find that negative text prompts provide substantial improvements: on the NC-setting, adding $T_{\text{neg}}$ reduces MAE from 32.22 to 26.67, and on the KC-setting from 15.96 to 13.22. 


\setlength{\tabcolsep}{3pt}
\begin{table}[t]
\centering
\begin{tabular}{cccccccc}
\toprule
\multicolumn{4}{c}{Input Modalities} & \multicolumn{2}{c}{NC-setting} & \multicolumn{2}{c}{KC-setting} \\
$T_{\text{pos}}$ & $E_{\text{pos}}$ & $T_{\text{neg}}$ & $E_{\text{neg}}$ & MAE & RMSE & MAE & RMSE \\
\midrule
\checkmark & & & & 32.22 & 48.38 & 15.96 & 28.76\\
\checkmark & \checkmark & & & 31.57 & 46.73 & 15.75 & 27.59 \\
\checkmark & & \checkmark & & 26.67 & 39.00 & 13.22 & 26.23\\
\checkmark & \checkmark & \checkmark & \checkmark & \textbf{26.61} & \textbf{38.86} & \textbf{12.72} & \textbf{23.99} \\
\bottomrule
\end{tabular}
\vskip -0.1in
\caption{Performance of different combinations of multimodal prompts on the test set of CoCount.}
\vskip -0.2in
\label{tab:ablation_modality}
\end{table}

\myheading{Using irrelevant negative prompt.} To further analyze the role of exclusion prompts, we conduct an additional experiment using irrelevant negative prompts that specify distractors not present in the scene. For instance, in a scene containing only black and brown candies where the goal is to count black ones, we replace the correct exclusion prompt (``not brown'') with an unrelated one (``not red''). Using such mismatched negatives, the model achieves an MAE of 28.43 and RMSE of 42.29 in the NC-setting, worse than using relevant negatives (MAE 26.67) but still better than using no exclusion prompt at all (MAE 32.22). This result highlights two points: (1) relevant exclusion prompts are crucial for optimal performance, and (2) even irrelevant exclusion cues can offer modest benefits due to CountEx's discriminative query embedding and selection mechanisms.




\setlength{\tabcolsep}{3pt}
\begin{table}[t]
\centering
\begin{tabular}{cccccc}
\toprule
\multirow{2}{*}{$\mathcal{L}_{\text{den}}$} & \multirow{2}{*}{$\mathcal{L}_{\text{proto}}$} & \multicolumn{2}{c}{NC-setting} & \multicolumn{2}{c}{KC-setting} \\
& & MAE & RMSE & MAE & RMSE \\
\midrule
 & & 30.75  & 42.84 & 15.20 & 28.33 \\
\checkmark &  & 29.21 & 41.61 & 13.00 & 25.22 \\
\checkmark & \checkmark & \textbf{26.61} & \textbf{38.86} & \textbf{12.72} & \textbf{23.99} \\
\bottomrule
\end{tabular}
\vskip -0.1in
\caption{Ablation on loss components. Both density prediction loss ($\mathcal{L}_{\text{den}}$) and the proto losses ($\mathcal{L}_{\text{proto}}$) contribute to improved performance. }
\label{tab:ablation_loss}
\vskip -0.3in
\end{table}

\myheading{Impact of Loss Components.} \Tref{tab:ablation_loss} evaluates the contribution of different loss components. Adding density prediction loss $\mathcal{L}_{\text{den}}$ improves NC-setting MAE from 30.75 to 29.21, demonstrating that dense supervision helps the encoder learn spatially-aware representations. Incorporating the proto losses $\mathcal{L}_{\text{proto}}$ (shareability and compactness) further reduces MAE to 26.61, showing that explicit guidance for learning shared features and negative-exclusive patterns is crucial for discriminative query refinement.



\subsection{Qualitative results and limitations}

\begin{figure}[t]
\centering
\begin{subfigure}[b]{0.45\textwidth}
    \centering
    \includegraphics[width=\textwidth]{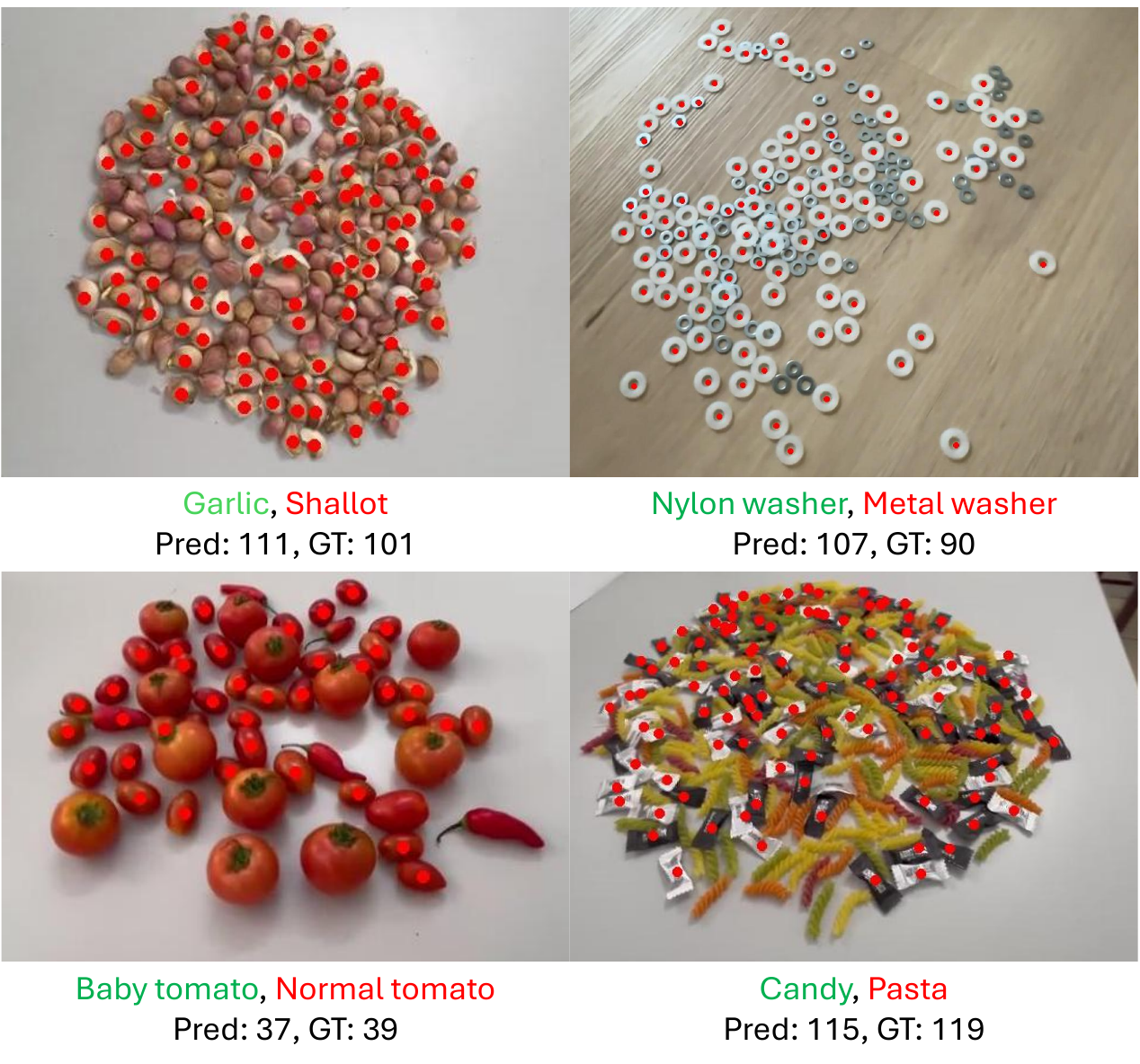}
    \vskip -0.05in
    \caption{General qualitative results across diverse categories}
    \label{fig:qual_general}
\end{subfigure}
\hfill
\begin{subfigure}[b]{0.45\textwidth}
    \centering
    \includegraphics[width=\textwidth]{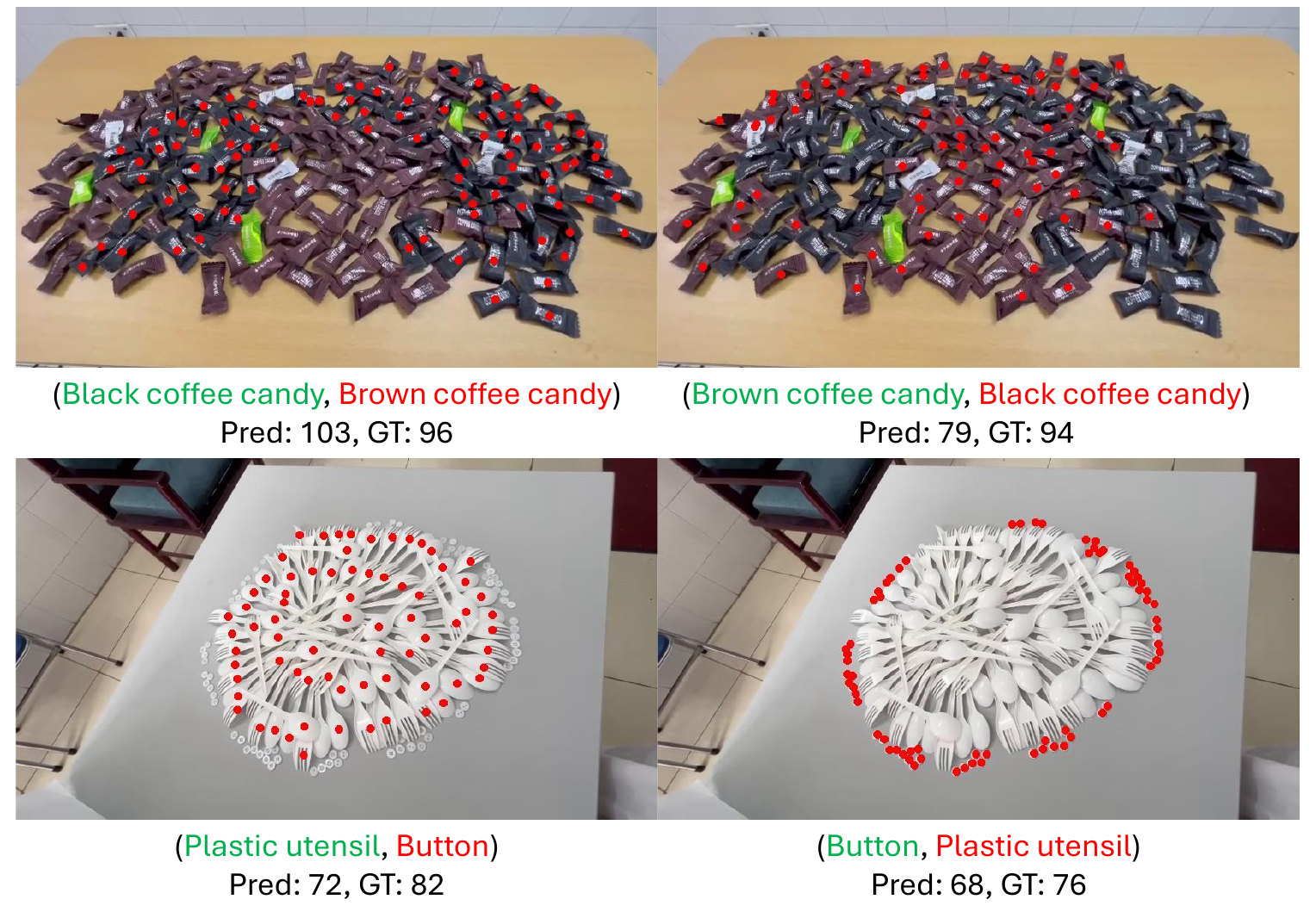}
    \vskip -0.05in
    \caption{Effect of swapping positive and negative prompts}
    \label{fig:qual_swap}
\end{subfigure}
\vskip -0.1in
\caption{Qualitative results on CoCount test set.}
\vskip -0.2in
\label{fig:qualitative_results}
\end{figure}

\Fref{fig:qualitative_results} illustrates CountEx's qualitative performance. Figure~\ref{fig:qual_general} demonstrates accurate counting across visually similar categories. Figure~\ref{fig:qual_swap} shows results on identical images when swapping positive and negative text prompts.

\begin{figure}[t]
    \centering
    \includegraphics[width=0.4\textwidth]{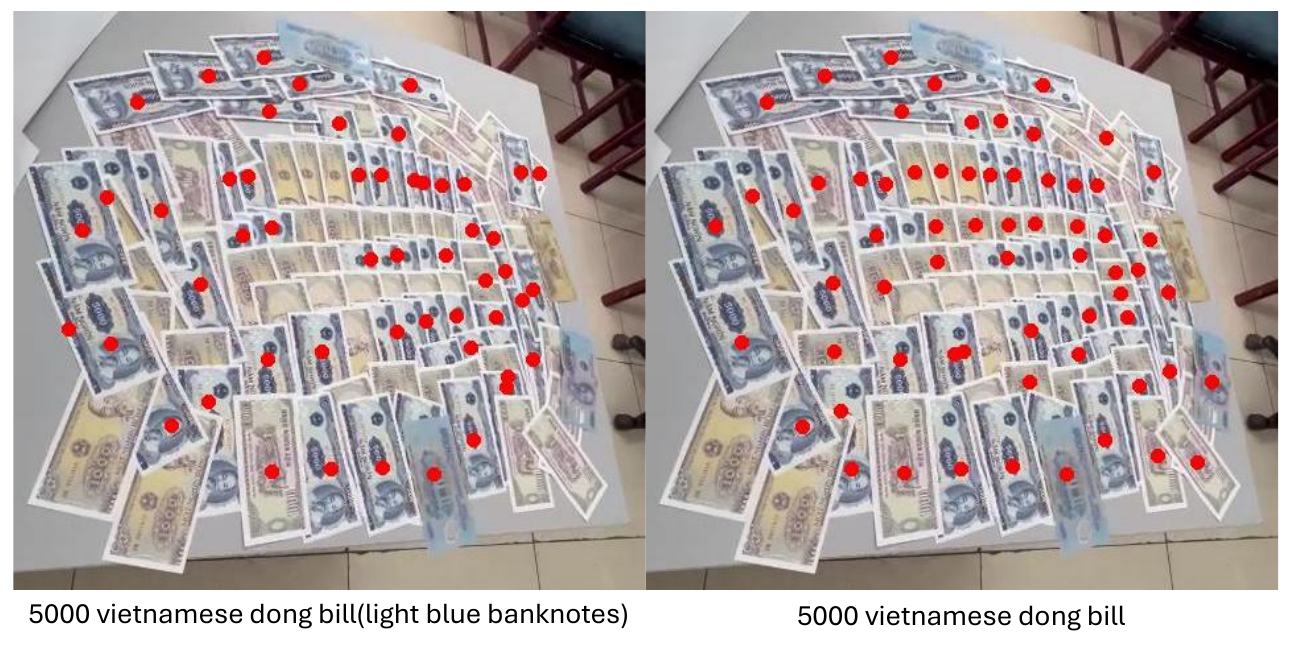}
    \vskip -0.1in
    \caption{Impact of prompt specificity.}
    \vskip -0.3in
    \label{fig:failure_cases}
\end{figure}

While CountEx achieves strong performance, two limitations remain. First, the model exhibits positive text dominance due to fine-tuning from a text-centric detector (LLMDet), causing over-reliance on textual descriptions. Second, BERT's limited reasoning capability restricts performance on vague prompts. Figure~\ref{fig:failure_cases} illustrates this: the generic prompt (5000 Vietnamese dong bill) yields poor accuracy, while adding explicit visual details (light blue banknotes) substantially improves results. Unlike large language models, BERT cannot infer discriminative visual features from abstract descriptions.


\section{Conclusions}

We introduced \textbf{CountEx}, a discriminative counting framework that leverages multimodal prompts to specify both inclusion and exclusion intent, enabling accurate fine-grained object counting.  To support rigorous evaluation, we presented \textbf{CoCount}, a benchmark with dense annotations across 97 fine-grained category pairs. Extensive experiments demonstrate that CountEx achieves state-of-the-art performance on CoCount, and generalizes effectively to LOOKALIKES, PairTally, and FSC-147. 

\myheading{Acknowledgment.} This work was funded by the Australian Institute for Machine Learning (Adelaide University) and the Centre for Augmented Reasoning, an initiative by the Department of Education, Australian Government. The authors would also like to
thank Viet Bach Tran for assistance with data collection.




{
    \small
    \bibliographystyle{ieeenat_fullname}
    \bibliography{main}

@String(CVPR= {IEEE Conf. Comput. Vis. Pattern Recog.})

@String(ICCV= {Int. Conf. Comput. Vis.})

@String(ECCV= {Eur. Conf. Comput. Vis.})

@String(NIPS= {Adv. Neural Inform. Process. Syst.})

@String(ACMMM= {ACM Int. Conf. Multimedia})

@String(ACCV  = {ACCV})

@String(AAAI = {AAAI})

@String(CVPR  = {CVPR})

@String(ICCV  = {ICCV})

@String(ECCV  = {ECCV})

@String(NIPS  = {NeurIPS})

@String(ACMMM = {ACM MM})

@String(WACV  = {WACV})

@inproceedings{m_Ranjan-etal-ECCV18,
author = {Viresh Ranjan and Hieu Le and Minh Hoai},
title = {Iterative Crowd Counting},
year = {2018},
booktitle = eccv
}

@inproceedings{m_Huang-etal-ICCV23,
 author = {Yifeng Huang and Viresh Ranjan and Minh Hoai},
 title = {Interactive Class-Agnostic Object Counting},
 year = {2023},
 booktitle = iccv,
}

@inproceedings{llmdet,
  title={Llmdet: Learning strong open-vocabulary object detectors under the supervision of large language models},
  author={Fu, Shenghao and Yang, Qize and Mo, Qijie and Yan, Junkai and Wei, Xihan and Meng, Jingke and Xie, Xiaohua and Zheng, Wei-Shi},
  booktitle=CVPR,
  year={2025}
}

@article{countgd,
  title={Countgd: Multi-modal open-world counting},
  author={Amini-Naieni, Niki and Han, Tengda and Zisserman, Andrew},
  journal=NIPS,
  year={2024}
}

@inproceedings{nguyen2022few,
  title={Few-shot Object Counting and Detection},
  author={Nguyen, Thanh and Pham, Chau and Nguyen, Khoi and Hoai, Minh},
  booktitle=ECCV,
  year={2022},
}

@inproceedings{shi2022represent,
  title={Represent, Compare, and Learn: A Similarity-Aware Framework for Class-Agnostic Counting},
  author={Shi, Min and Lu, Hao and Feng, Chen and Liu, Chengxin and Cao, Zhiguo},
  booktitle=CVPR,
  year={2022}
}

@inproceedings{Zhiyuan2023safecount,
  title={Few-shot Object Counting with Similarity-Aware Feature Enhancement},
  author={You, Zhiyuan and Yang, Kai and Luo, Wenhan and Lu, Xin and Cui, Lei and Le, Xinyi},
  booktitle=WACV,
  year={2023}
}

@inproceedings{wang2020DMCount,
  title={Distribution Matching for Crowd Counting},
  author={Boyu Wang and Huidong Liu and Dimitris Samaras and Minh Hoai},
  booktitle=NIPS,
  year={2020},
}

@inproceedings{ranjan2021learning,
  title={Learning To Count Everything},
  author={Ranjan, Viresh and Sharma, Udbhav and Nguyen, Thu and Hoai, Minh},
  booktitle=CVPR,
  year={2021}
}

@inproceedings{xu2019learn,
  title={Learn to scale: Generating multipolar normalized density maps for crowd counting},
  author={Xu, Chenfeng and Qiu, Kai and Fu, Jianlong and Bai, Song and Xu, Yongchao and Bai, Xiang},
  booktitle=ICCV,
  year={2019}
}

@inproceedings{song2021rethinking,
  title={Rethinking Counting and Localization in Crowds: A Purely Point-Based Framework},
  author={Song, Qingyu and Wang, Changan and Jiang, Zhengkai and Wang, Yabiao and Tai, Ying and Wang, Chengjie and Li, Jilin and Huang, Feiyue and Wu, Yang},
  booktitle=ICCV,
  year={2021}
}

@inproceedings{liu2019point,
  title={Point in, box out: Beyond counting persons in crowds},
  author={Liu, Yuting and Shi, Miaojing and Zhao, Qijun and Wang, Xiaofang},
  booktitle=CVPR,
  year={2019}
}

@inproceedings{hu2020count,
  title={Nas-count: Counting-by-density with neural architecture search},
  author={Hu, Yutao and Jiang, Xiaolong and Liu, Xuhui and Zhang, Baochang and Han, Jungong and Cao, Xianbin and Doermann, David},
  booktitle=ECCV,
  year={2020},
}

@inproceedings{li2018csrnet,
  title={Csrnet: Dilated convolutional neural networks for understanding the highly congested scenes},
  author={Li, Yuhong and Zhang, Xiaofan and Chen, Deming},
  booktitle=CVPR,
  year={2018}
}

@inproceedings{miao2020shallow,
  title={Shallow feature based dense attention network for crowd counting},
  author={Miao, Yunqi and Lin, Zijia and Ding, Guiguang and Han, Jungong},
  booktitle=AAAI,
  year={2020}
}

@inproceedings{jiang2020attention,
  title={Attention scaling for crowd counting},
  author={Jiang, Xiaoheng and Zhang, Li and Xu, Mingliang and Zhang, Tianzhu and Lv, Pei and Zhou, Bing and Yang, Xin and Pang, Yanwei},
  booktitle=CVPR,
  year={2020}
}

@inproceedings{bai2020adaptive,
  title={Adaptive dilated network with self-correction supervision for counting},
  author={Bai, Shuai and He, Zhiqun and Qiao, Yu and Hu, Hanzhe and Wu, Wei and Yan, Junjie},
  booktitle=CVPR,
  year={2020}
}

@inproceedings{liu2020adaptive,
  title={Adaptive mixture regression network with local counting map for crowd counting},
  author={Liu, Xiyang and Yang, Jie and Ding, Wenrui and Wang, Tieqiang and Wang, Zhijin and Xiong, Junjun},
  booktitle={ECCV},
  year={2020},
}

@inproceedings{lian2019density,
  title={Density map regression guided detection network for rgb-d crowd counting and localization},
  author={Lian, Dongze and Li, Jing and Zheng, Jia and Luo, Weixin and Gao, Shenghua},
  booktitle=CVPR,
  year={2019}
}

@article{sam2020locate,
  title={Locate, size and count: Accurately resolving people in dense crowds via detection},
  author={Sam, Deepak Babu and Peri, Skand Vishwanath and Sundararaman, Mukuntha Narayanan and Kamath, Amogh and Radhakrishnan, Venkatesh Babu},
  journal={IEEE transactions on pattern analysis and machine intelligence},
  year={2020},
  publisher={IEEE}
}

@inproceedings{liu2019recurrent,
  title={Recurrent attentive zooming for joint crowd counting and precise localization},
  author={Liu, Chenchen and Weng, Xinyu and Mu, Yadong},
  booktitle=CVPR,
  year={2019}
}

@inproceedings{laradji2018blobs,
  title={Where are the blobs: Counting by localization with point supervision},
  author={Laradji, Issam H and Rostamzadeh, Negar and Pinheiro, Pedro O and Vazquez, David and Schmidt, Mark},
  booktitle={ECCV},
  year={2018}
}

@inproceedings{zhang2016single,
  title={Single-image crowd counting via multi-column convolutional neural network},
  author={Zhang, Yingying and Zhou, Desen and Chen, Siqin and Gao, Shenghua and Ma, Yi},
  booktitle={Proceedings of the IEEE conference on computer vision and pattern recognition},
  year={2016}
}

@inproceedings{yang2021class,
  title={Class-agnostic few-shot object counting},
  author={Yang, Shuo-Diao and Su, Hung-Ting and Hsu, Winston H and Chen, Wen-Chin},
  booktitle=WACV,
  year={2021}
}

@inproceedings{lu2018class,
  title={Class-agnostic counting},
  author={Lu, Erika and Xie, Weidi and Zisserman, Andrew},
  booktitle=ACCV,
  year={2018},
}

@inproceedings{liu2024grounding,
  title={Grounding dino: Marrying dino with grounded pre-training for open-set object detection},
  author={Liu, Shilong and Zeng, Zhaoyang and Ren, Tianhe and Li, Feng and Zhang, Hao and Yang, Jie and Jiang, Qing and Li, Chunyuan and Yang, Jianwei and Su, Hang and others},
  booktitle={ECCV},
  year={2024},
}

@inproceedings{pelhan2024dave,
  title={Dave-a detect-and-verify paradigm for low-shot counting},
  author={Pelhan, Jer and Zavrtanik, Vitjan and Kristan, Matej and others},
  booktitle=CVPR,
  year={2024}
}

@inproceedings{djukic2023loca,
  title={A low-shot object counting network with iterative prototype adaptation},
  author={{DJ}uki{\'c}, Nikola and Luke{\v{z}}i{\v{c}}, Alan and Zavrtanik, Vitjan and Kristan, Matej},
  booktitle=CVPR,
  year={2023}
}

@article{pelhan2024novel,
  title={A novel unified architecture for low-shot counting by detection and segmentation},
  author={Pelhan, Jer and Lukezic, Alan and Zavrtanik, Vitjan and Kristan, Matej},
  journal=NIPS,
  year={2024}
}

@article{lu2024ovis,
  title={Ovis: Structural embedding alignment for multimodal large language model},
  author={Lu, Shiyin and Li, Yang and Chen, Qing-Guo and Xu, Zhao and Luo, Weihua and Zhang, Kaifu and Ye, Han-Jia},
  journal={arXiv:2405.20797},
  year={2024}
}

@inproceedings{ranjan2022exemplar,
  title={Exemplar free class agnostic counting},
  author={Ranjan, Viresh and Nguyen, Minh Hoai},
  booktitle=ACCV,
  year={2022}
}

@article{liu2022countr,
  title={Countr: Transformer-based generalised visual counting},
  author={Liu, Chang and Zhong, Yujie and Zisserman, Andrew and Xie, Weidi},
  journal={arXiv preprint arXiv:2208.13721},
  year={2022}
}

@article{hobley2022learning,
  title={Learning to count anything: Reference-less class-agnostic counting with weak supervision},
  author={Hobley, Michael and Prisacariu, Victor},
  journal={arXiv preprint arXiv:2205.10203},
  year={2022}
}

@inproceedings{jiang2023clip,
  title={Clip-count: Towards text-guided zero-shot object counting},
  author={Jiang, Ruixiang and Liu, Lingbo and Chen, Changwen},
  booktitle=ACMMM,
  year={2023}
}

@inproceedings{kang2024vlcounter,
  title={Vlcounter: Text-aware visual representation for zero-shot object counting},
  author={Kang, Seunggu and Moon, WonJun and Kim, Euiyeon and Heo, Jae-Pil},
  booktitle=AAAI,
  year={2024}
}

@article{amini2023open,
  title={Open-world text-specified object counting},
  author={Amini-Naieni, Niki and Amini-Naieni, Kiana and Han, Tengda and Zisserman, Andrew},
  journal={arXiv:2306.01851},
  year={2023}
}

@inproceedings{xu2023zero,
  title={Zero-shot object counting},
  author={Xu, Jingyi and Le, Hieu and Nguyen, Vu and Ranjan, Viresh and Samaras, Dimitris},
  booktitle=CVPR,
  year={2023}
}

@inproceedings{zhu2024zero,
  title={Zero-shot object counting with good exemplars},
  author={Zhu, Huilin and Yuan, Jingling and Yang, Zhengwei and Guo, Yu and Wang, Zheng and Zhong, Xian and He, Shengfeng},
  booktitle={ECCV},
  year={2024},  
}

@inproceedings{wang2024vision,
  title={Vision transformer off-the-shelf: A surprising baseline for few-shot class-agnostic counting},
  author={Wang, Zhicheng and Xiao, Liwen and Cao, Zhiguo and Lu, Hao},
  booktitle=AAAI,
  year={2024}
}

@inproceedings{radford2021learning,
  title={Learning Transferable Visual Models from Natural Language Supervision},
  author={Radford, Alec and Kim, Jong Wook and Hallacy, Chris and Ramesh, Aditya and Goh, Gabriel and Agarwal, Sandhini and Sastry, Girish and Askell, Amanda and Mishkin, Pamela and Clark, Jack and others},
  booktitle={International Conference on Machine Learning},
  year={2021},
}

@inproceedings{Dai_2024_CVPR,
  title={Referring Expression Counting},
  author={Dai, Siyang and Liu, Jun and Cheung, Ngai-Man},
  booktitle=CVPR,
  year={2024}
}

@inproceedings{huang2024point,
  title={Point segment and count: A generalized framework for object counting},
  author={Huang, Zhizhong and Dai, Mingliang and Zhang, Yi and Zhang, Junping and Shan, Hongming},
  booktitle=CVPR,
  year={2024}
}

@inproceedings{shi2024training,
  title={Training-free object counting with prompts},
  author={Shi, Zenglin and Sun, Ying and Zhang, Mengmi},
  booktitle=WACV,
  year={2024}
}

@inproceedings{zhang2025enhancing,
  title={Enhancing Zero-shot Object Counting via Text-guided Local Ranking and Number-evoked Global Attention},
  author={Zhang, Shiwei and Zhou, Qi and Ke, Wei},
  booktitle=ICCV,
  year={2025}
}

@inproceedings{liu2025countse,
  title={CountSE: Soft Exemplar Open-set Object Counting},
  author={Liu, Shuai and2025justd Zhang, Peng and Zhang, Shiwei and Ke, Wei},
  booktitle=ICCV,
  year={2025}
}

@inproceedings{wang2025exploring,
  title={Exploring Contextual Attribute Density in Referring Expression Counting},
  author={Wang, Zhicheng and Pan, Zhiyu and Peng, Zhan and Cheng, Jian and Xiao, Liwen and Jiang, Wei and Cao, Zhiguo},
  booktitle=CVPR,
  year={2025}
}

@article{d2025just,
  title={Just Say the Word: Annotation-Free Fine-Grained Object Counting},
  author={D'Alessandro, Adriano and Mahdavi-Amiri, Ali and Hamarneh, Ghassan},
  journal={arXiv preprint arXiv:2504.11705},
  year={2025}
}

@article{nguyen2025can,
  title={Can Current AI Models Count What We Mean, Not What They See? A Benchmark and Systematic Evaluation},
  author={Nguyen, Gia Khanh and Huang, Yifeng and Hoai, Minh},
  journal={arXiv preprint arXiv:2509.13939},
  year={2025}
}

@inproceedings{karaev2025cotracker3,
  title={Cotracker3: Simpler and better point tracking by pseudo-labelling real videos},
  author={Karaev, Nikita and Makarov, Yuri and Wang, Jianyuan and Neverova, Natalia and Vedaldi, Andrea and Rupprecht, Christian},
  booktitle=ICCV,
  year={2025}
}

@article{minderer2023scaling,
  title={Scaling open-vocabulary object detection},
  author={Minderer, Matthias and Gritsenko, Alexey and Houlsby, Neil},
  journal=NIPS,
  year={2023}
}

@article{Qwen2.5-VL,
  title={Qwen2. 5-vl technical report},
  author={Bai, Shuai and Chen, Keqin and Liu, Xuejing and Wang, Jialin and Ge, Wenbin and Song, Sibo and Dang, Kai and Wang, Peng and Wang, Shijie and Tang, Jun and others},
  journal={arXiv:2502.13923},
  year={2025}
}

@article{zhu2025internvl3exploringadvancedtraining,
  title={Internvl3: Exploring advanced training and test-time recipes for open-source multimodal models},
  author={Zhu, Jinguo and Wang, Weiyun and Chen, Zhe and Liu, Zhaoyang and Ye, Shenglong and Gu, Lixin and Tian, Hao and Duan, Yuchen and Su, Weijie and Shao, Jie and others},
  journal={arXiv:2504.10479},
  year={2025}
}

@article{dubey2024llama,
  title={The llama 3 herd of models},
  author={Dubey, Abhimanyu and Jauhri, Abhinav and Pandey, Abhinav and Kadian, Abhishek and Al-Dahle, Ahmad and Letman, Aiesha and Mathur, Akhil and Schelten, Alan and Yang, Amy and Fan, Angela and others},
  journal={arXiv--2407},
  year={2024}
}
}
\clearpage
\setcounter{page}{1}
\maketitlesupplementary

\section{Overview}
This supplementary material contains additional experimental results, dataset statistics, and qualitative examples omitted from the main paper. We provide complete validation and test results on CoCount (KC-setting and all five NC-setting splits), detailed FSC-147 results, object count distribution, and additional qualitative visualizations.

\section{Additional Experiment Result}

This section provides complete experimental results omitted from the main paper due to space constraints, including validation results on NC-setting and KC-setting, per-supercategory NC-setting breakdowns, and detailed FSC-147 results.

\subsection{Complete Results on CoCount}

\subsubsection{Known-Category (KC) Setting}
\Tref{tab:kc_setting_exp} presents complete validation and test results on the KC-setting. CountEx achieves the best performance on both splits, with 14.07 validation MAE and 12.72 test MAE.

\begin{table}[!ht]
\centering
\begin{tabular}{lcccc}
\toprule
\multirow{2}{*}{Method} & \multicolumn{2}{c}{Val} & \multicolumn{2}{c}{Test} \\
& MAE & RMSE & MAE & RMSE \\
\midrule
LLMDet~\cite{llmdet} & 17.60 & 30.50 & 16.82 & 29.23 \\
CAD-GD~\cite{wang2025exploring} & 15.85 & 26.92 & 16.00 & 27.52 \\
GroundingREC~\cite{Dai_2024_CVPR} & 17.72 & 27.83 & 17.54 & 27.41 \\
CountGD~\cite{countgd} & 16.06 & 28.39 & 15.55 & 28.32 \\
CountEx (proposed) & \textbf{14.07} & \textbf{27.61} & \textbf{12.72} & \textbf{23.99} \\
\bottomrule
\end{tabular}
\caption{Complete results on CoCount KC-setting.}
\label{tab:kc_setting_exp}
\end{table}

\subsubsection{Novel-Category (NC) Setting}

Tables~\ref{tab:nc_setting_exp_foo}--\ref{tab:nc_setting_exp_fun} present complete results for all five NC-setting splits, where each supercategory serves as the held-out test set.

\begin{table}[!ht]
\centering
\begin{tabular}{lcccc}
\toprule
\multirow{2}{*}{Method} & \multicolumn{2}{c}{Val} & \multicolumn{2}{c}{Test} \\
& MAE & RMSE & MAE & RMSE \\
\midrule
LLMDet~\cite{llmdet} & 45.87 & 65.71 & 45.92 & 65.69 \\
CAD-GD~\cite{wang2025exploring} & 56.52 & 81.25 & 56.94 & 81.64 \\
GroundingREC~\cite{Dai_2024_CVPR} & 41.24 & 58.28 & 40.85 & 57.22 \\
CountGD~\cite{countgd} & 44.00 & 60.17 & 41.63 & 57.50 \\
CountEx (proposed) & \textbf{39.41} & \textbf{54.82} & \textbf{37.04} & \textbf{50.58} \\
\bottomrule
\end{tabular}
\caption{Complete results on NC-setting Test-Food.}
\label{tab:nc_setting_exp_foo}
\end{table}

\begin{table}[!ht]
\centering
\begin{tabular}{lcccc}
\toprule
\multirow{2}{*}{Method} & \multicolumn{2}{c}{Val} & \multicolumn{2}{c}{Test} \\
& MAE & RMSE & MAE & RMSE \\
\midrule
LLMDet~\cite{llmdet} & 27.75 & 41.64 & 27.77 & 41.61 \\
CAD-GD~\cite{wang2025exploring} & \textbf{24.57} & \textbf{37.05} & 26.36 & 39.96 \\
GroundingREC~\cite{Dai_2024_CVPR} & 28.26 & 41.80 & 29.31 & 43.55\\
CountGD~\cite{countgd} & 35.06 & 48.92 & 32.73 & 47.24 \\
CountEx (proposed) & 25.55 & 37.72 & \textbf{24.16} & \textbf{34.87} \\
\bottomrule
\end{tabular}
\caption{Complete results on NC-setting Test-Home.}
\label{tab:nc_setting_exp_hou}
\end{table}

\begin{table}[!ht]
\centering
\begin{tabular}{lcccc}
\toprule
\multirow{2}{*}{Method} & \multicolumn{2}{c}{Val} & \multicolumn{2}{c}{Test} \\
& MAE & RMSE & MAE & RMSE \\
\midrule
LLMDet~\cite{llmdet} & 35.69 & 57.57 & 35.72 & 57.55\\
CAD-GD~\cite{wang2025exploring} & 39.28 & 57.29 & 36.28 & 54.47\\
GroundingREC~\cite{Dai_2024_CVPR} & 36.27 & 54.56 & 33.31 & \textbf{51.24}\\
CountGD~\cite{countgd} & 41.01 & 54.69 & 39.40 & 58.08 \\
CountEx (proposed) & \textbf{32.07} & \textbf{49.14} & \textbf{31.18} & 51.90 \\
\bottomrule
\end{tabular}
\caption{Complete results on NC-setting Test-Desk.}
\label{tab:nc_setting_exp_off}
\end{table}

\begin{table}[!ht]
\centering
\begin{tabular}{lcccc}
\toprule
\multirow{2}{*}{Method} & \multicolumn{2}{c}{Val} & \multicolumn{2}{c}{Test} \\
& MAE & RMSE & MAE & RMSE \\
\midrule
LLMDet~\cite{llmdet} & 32.51 & 42.56 & 32.39 & 42.44\\
CAD-GD~\cite{wang2025exploring} & 30.48 & 47.82 & 32.13 & 48.84\\
GroundingREC~\cite{Dai_2024_CVPR} & \textbf{26.49} & 39.80 & 25.04 & 34.38\\
CountGD~\cite{countgd} & 28.87 & 38.59 & 31.96 & 42.97 \\
CountEx (proposed) & 26.78 & \textbf{37.91} & \textbf{23.82} & \textbf{32.68}\\
\bottomrule
\end{tabular}
\caption{Complete results on NC-setting Test-Misc.}
\label{tab:nc_setting_exp_otr}
\end{table}

\begin{table}[!ht]
\centering
\begin{tabular}{lcccc}
\toprule
\multirow{2}{*}{Method} & \multicolumn{2}{c}{Val} & \multicolumn{2}{c}{Test} \\
& MAE & RMSE & MAE & RMSE \\
\midrule
LLMDet~\cite{llmdet} & 23.66 & 30.17 & 24.31 & 31.00\\
CAD-GD~\cite{wang2025exploring} & 19.64 & 28.10 & 18.67 & 27.02\\
GroundingREC~\cite{Dai_2024_CVPR} & 21.86 & 31.72 & 17.94 & 25.75\\
CountGD~\cite{countgd} & 20.92 & 31.44 & 23.18 & 35.68 \\
CountEx (proposed) & \textbf{16.64} & \textbf{23.50} & \textbf{16.84} & \textbf{24.26}\\
\bottomrule
\end{tabular}
\caption{Complete results on NC-setting Test-Game.}
\label{tab:nc_setting_exp_fun}
\end{table}

\subsection{Detailed Result on FSC-147}

Table~\ref{tab:fsc} presents complete results on FSC-147. We fine-tune our KC-setting model on the FSC-147 training set, following~\cite{countgd} that divides large images into patches and aggregates predictions. Positive prompts use captions from~\cite{amini2023open}, while negative prompts are set to ``None.''
\begin{table}[t]
\centering
\begin{tabular}{lccccc}
\toprule
\multirow{2}{*}{Method} & & \multicolumn{2}{c}{Val} & \multicolumn{2}{c}{Test} \\
\cmidrule(lr){3-4} \cmidrule(lr){5-6}
& & MAE & RMSE & MAE & RMSE \\
\midrule
CounTR~\cite{liu2022countr} & & 13.13 & 49.83 & 11.95 & 91.23 \\
LOCA~\cite{djukic2023loca} & & 10.24 & 32.56 & 10.79 & 56.97 \\
CACViT~\cite{wang2024vision} & & 10.63 & 37.95 & 9.13 & 48.96 \\
DAVE~\cite{pelhan2024dave} & & 8.91 & 28.08 & 8.66 & 32.36 \\
CountGD~\cite{countgd} & & 7.10 & 26.08 & 5.74 & 24.09 \\
\midrule
Patch-selection~\cite{xu2023zero} & & 26.93 & 88.63 & 22.09 & 115.17 \\
CLIP-count~\cite{jiang2023clip} & & 18.79 & 61.18 & 17.78 & 106.62 \\
VLCounter~\cite{kang2024vlcounter} & & 18.06 & 65.13 & 17.05 & 106.16 \\
CounTX~\cite{amini2023open} & & 17.10 & 65.61 & 15.88 & 106.29 \\
GroundingREC~\cite{Dai_2024_CVPR} & & 10.06 & 58.62 & 10.12 & 107.19 \\
CAD-GD~\cite{wang2025exploring} & & 9.30 & 40.96 & 10.35 & 86.88 \\
CountEx (proposed) & & 9.59 & 38.54 & 8.63 & 84.55 \\
\bottomrule
\end{tabular}
\caption{Complete results on FSC-147.}
\label{tab:fsc}
\end{table}

\section{Additional Dataset Statistics}
\begin{figure}[t]
  \centering
    \includegraphics[width=0.45\textwidth]{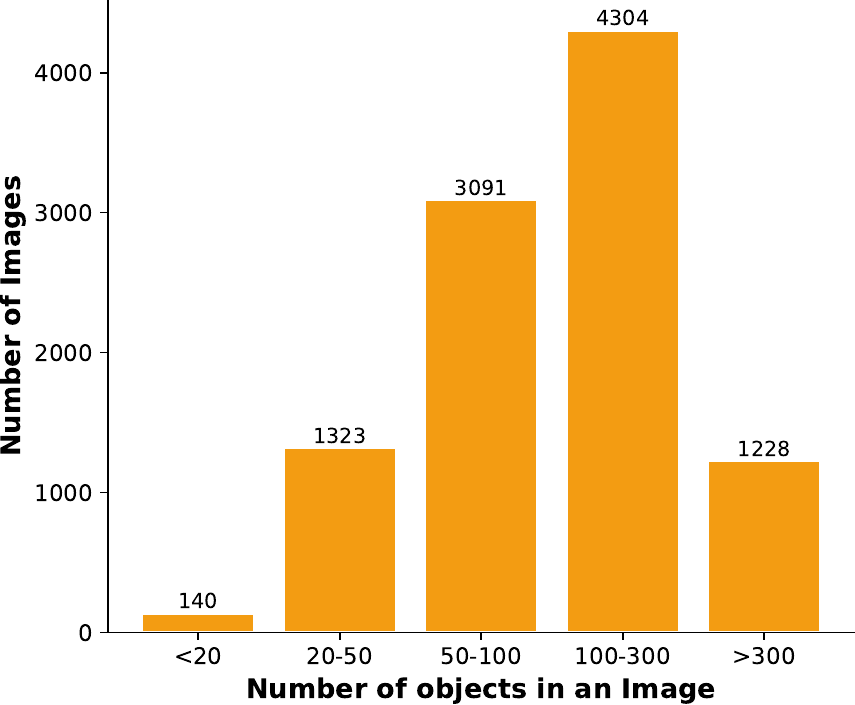}
    \caption{Object count distribution per image.} 
    \label{fig_sup:dataset_stats}
\end{figure}

\Fref{fig_sup:dataset_stats} shows the object count distribution per image in CoCount.

\section{Additional Qualitive Results}
This section presents additional qualitative results on the CoCount test set.

\subsection{Known-Category (KC) Setting}
\begin{figure}[t]
\centering
\includegraphics[width=0.45\textwidth]{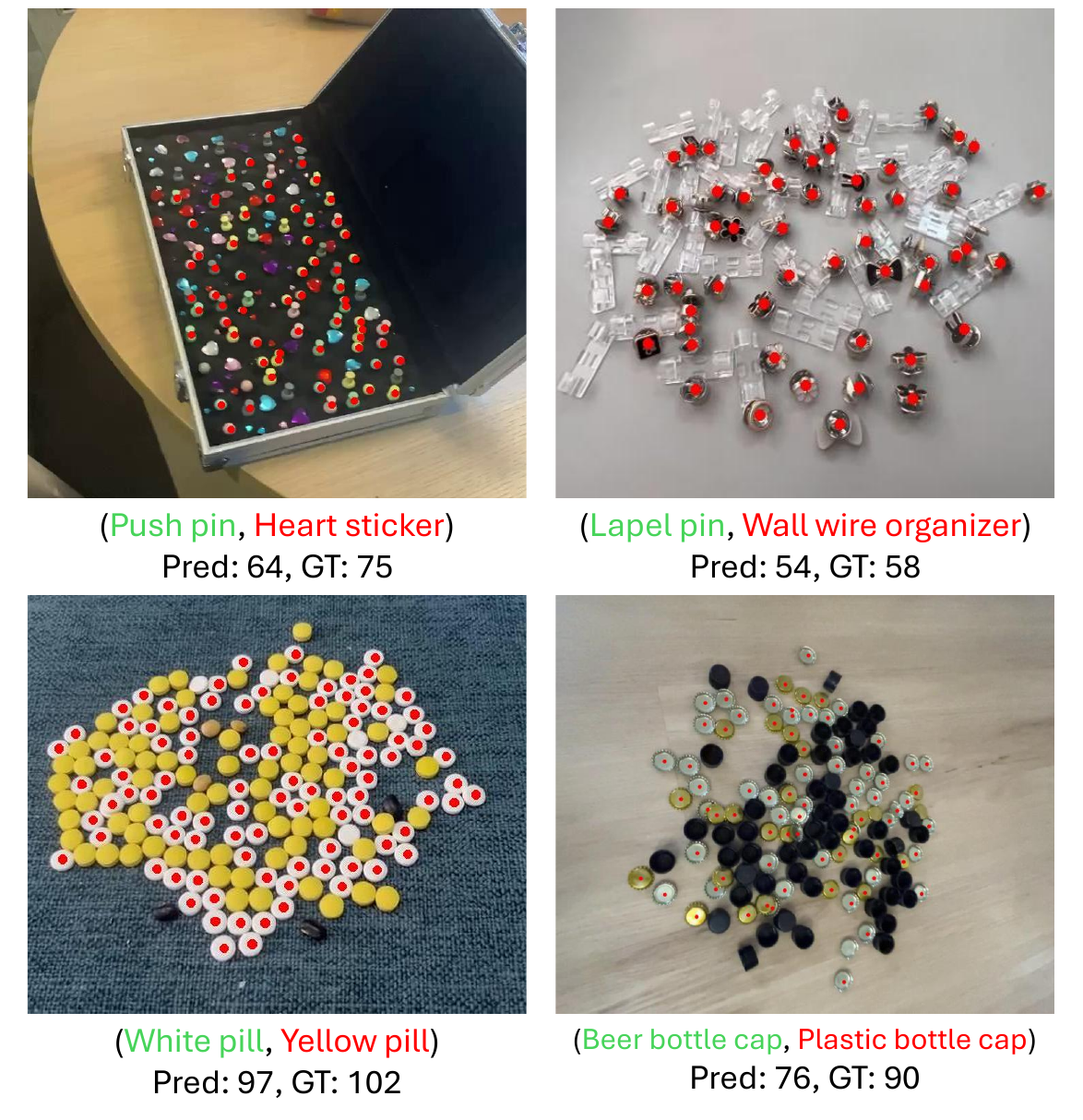}
\caption{Additional qualitative results on CoCount test set (KC-setting).}
\label{fig:qualitative_results_kc}
\end{figure}

Additional qualitative results on the CoCount test set under KC-setting are shown in~\Fref{fig:qualitative_results_kc}.

\subsection{Novel-Category (NC) Setting}

\begin{figure}[t]
\centering
\begin{subfigure}[b]{0.45\textwidth}
    \centering
    \includegraphics[width=\textwidth]{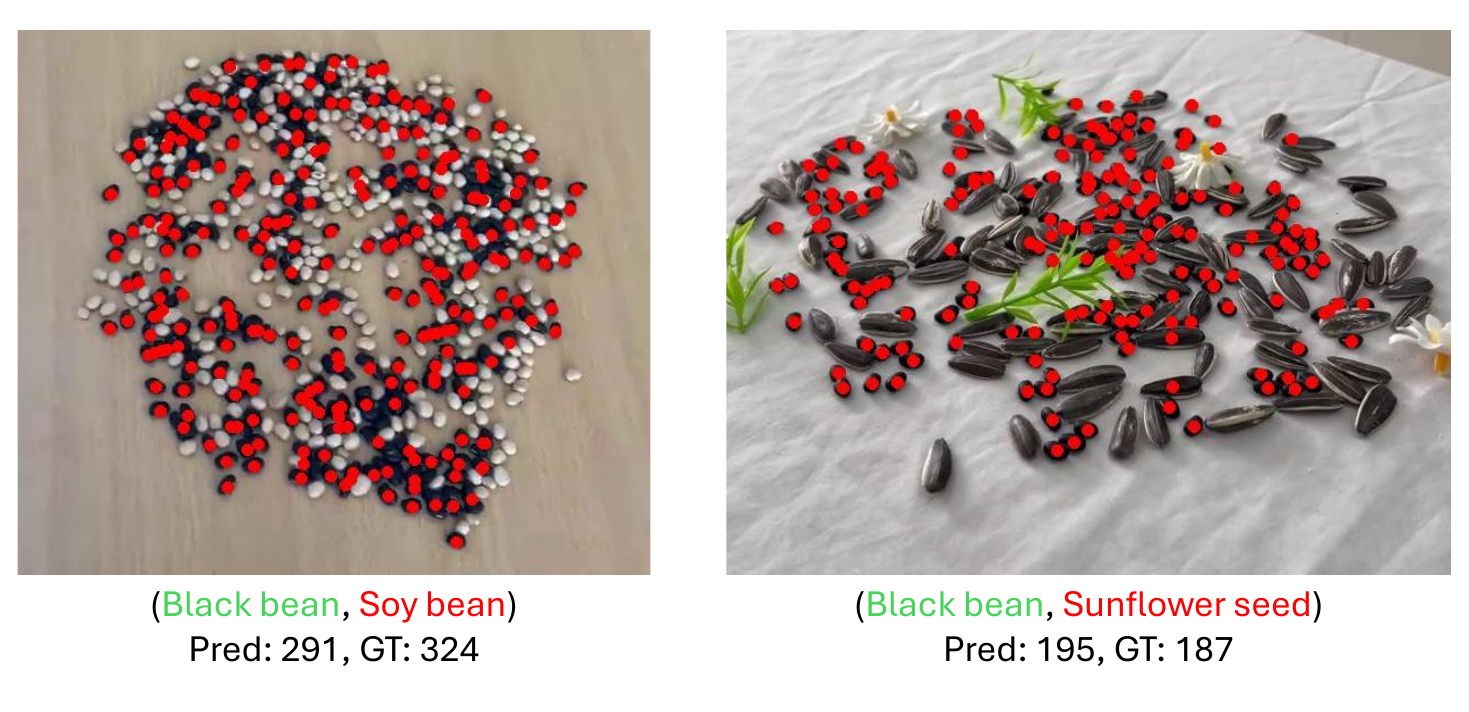}
    \caption{Test-Food.}
    \label{fig_sup:qual_general}
\end{subfigure}
\hfill
\begin{subfigure}[b]{0.45\textwidth}
    \centering
    \includegraphics[width=\textwidth]{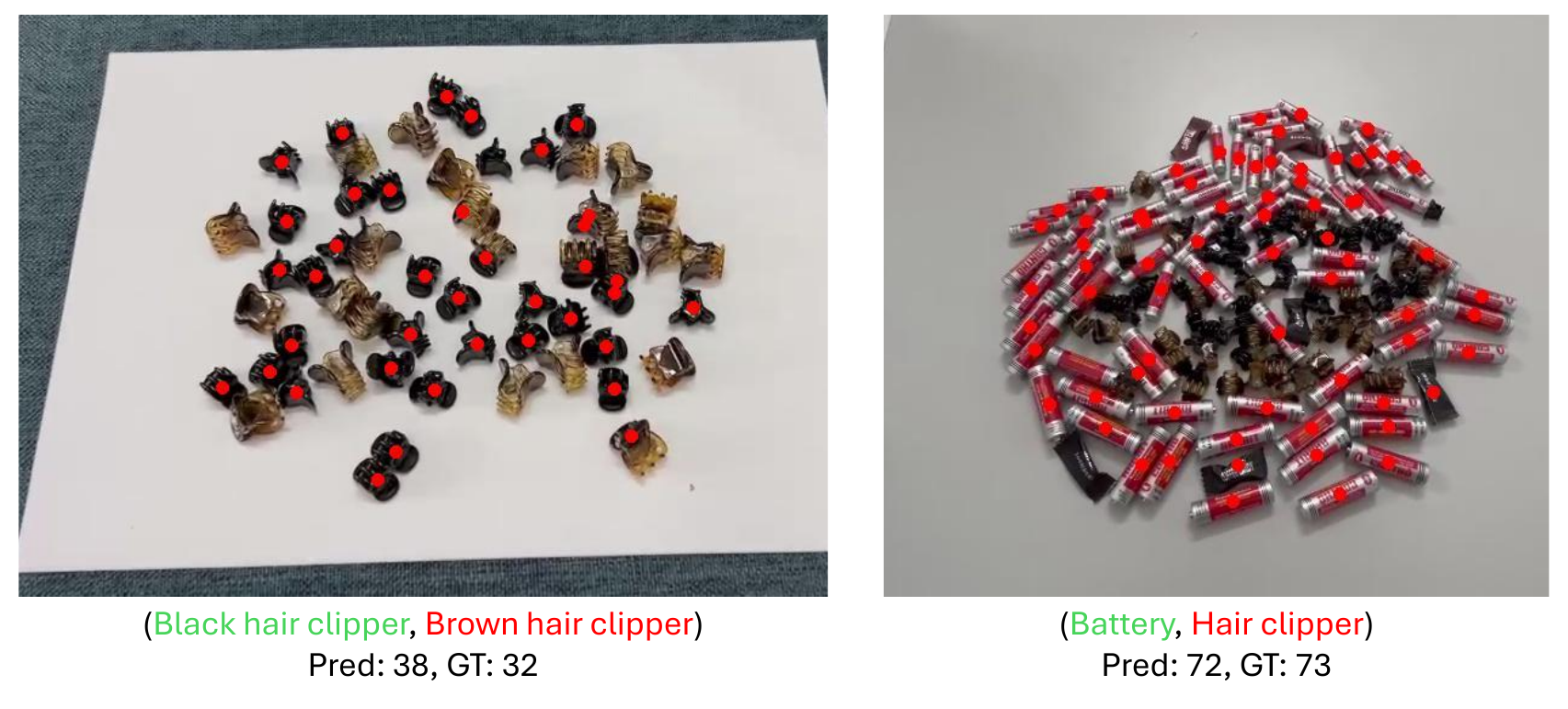}
    \caption{Test-Home.}
    \label{fig_sup:qual_home}
\end{subfigure}
\hfill
\begin{subfigure}[b]{0.45\textwidth}
    \centering
    \includegraphics[width=\textwidth]{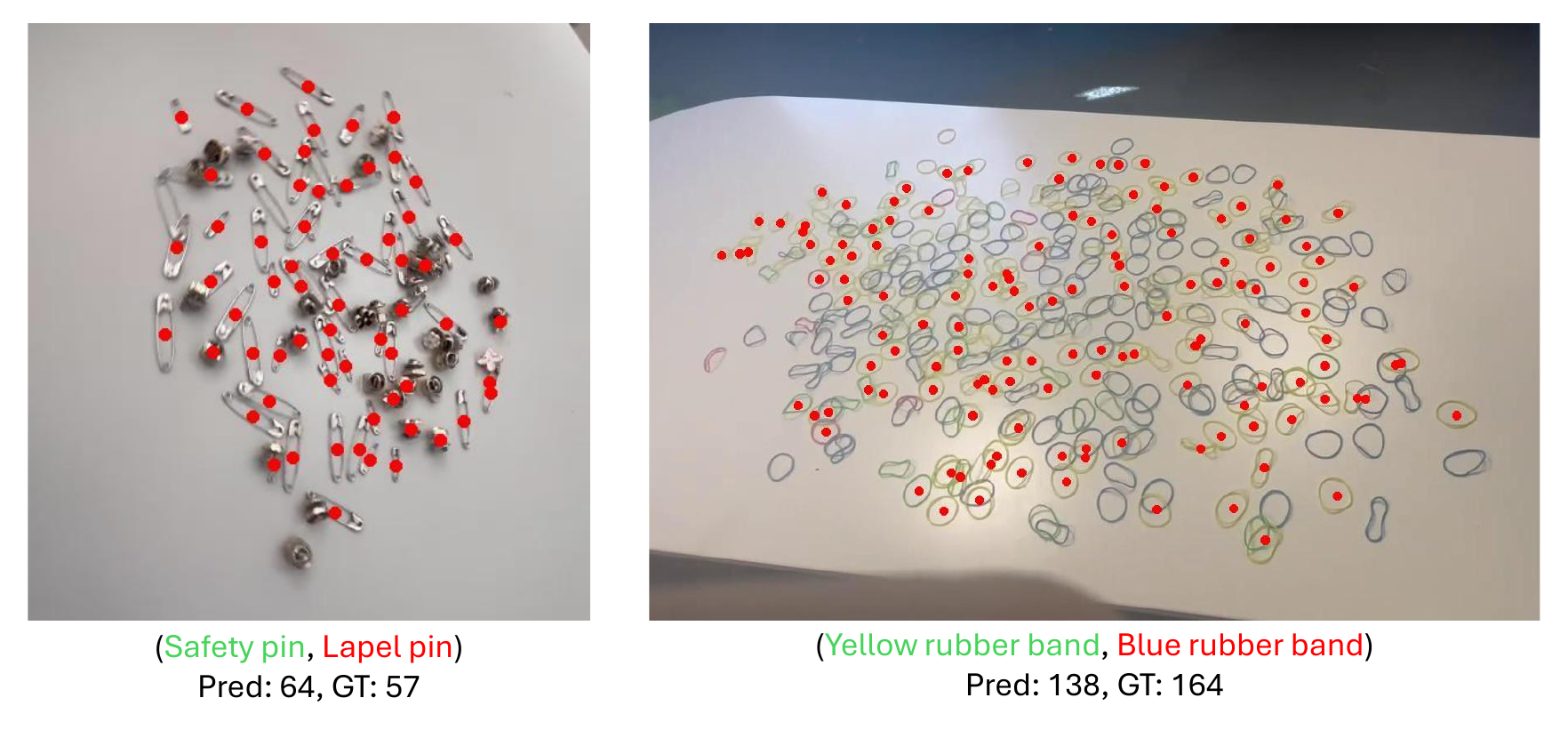}
    \caption{Test-Desk.}
    \label{fig_sup:qual_desk}
\end{subfigure}
\hfill
\begin{subfigure}[b]{0.45\textwidth}
    \centering
    \includegraphics[width=\textwidth]{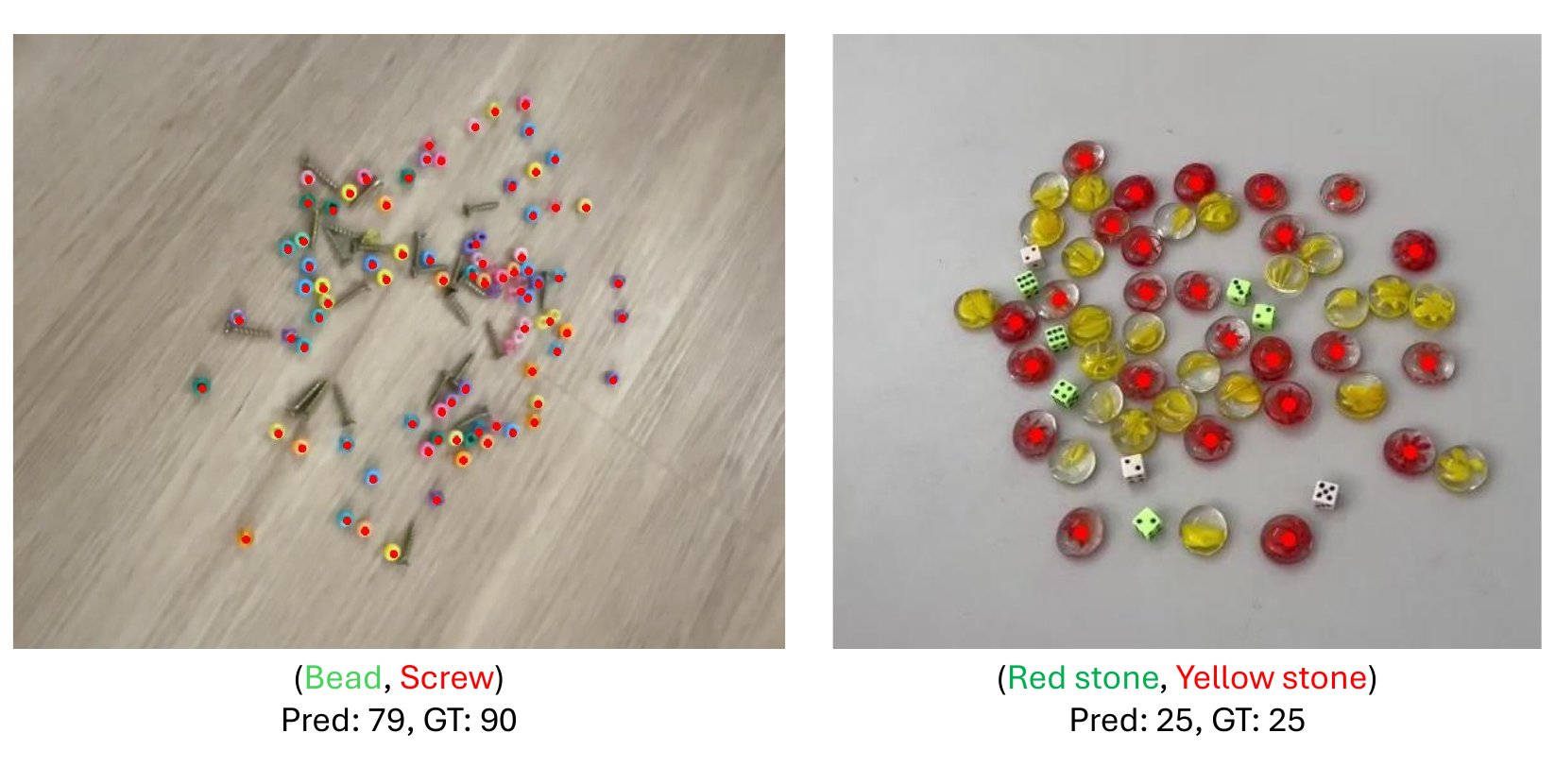}
    \caption{Test-Misc.}
    \label{fig_sup:qual_misc}
\end{subfigure}
\hfill
\begin{subfigure}[b]{0.45\textwidth}
    \centering
    \includegraphics[width=\textwidth]{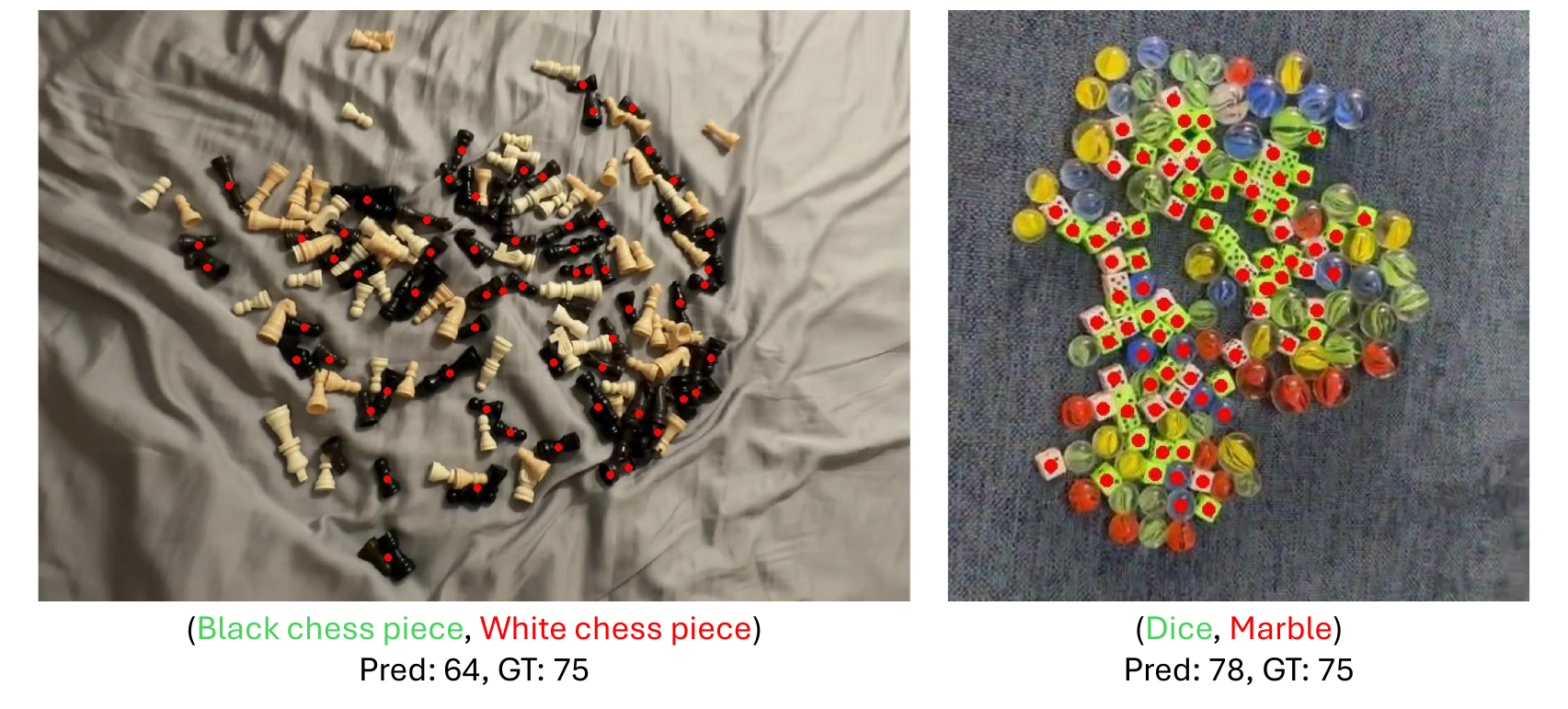}
    \caption{Test-Game.}
    \label{fig_sup:qual_game}
\end{subfigure}
\caption{Additional qualitative results on CoCount test set (NC-setting).}
\label{fig:qualitative_results_nc}
\end{figure}

\Fref{fig:qualitative_results_nc} shows additional qualitative results 
for all five NC-setting splits.

\end{document}